\documentclass[acmlarge]{acmart}

\usepackage{epigraph}
\usepackage{color}

\newcommand{\newtext}[1]{\textcolor{black}{#1}}

\usepackage[outercaption]{sidecap}   
\sidecaptionvpos{figure}{c}

\AtBeginDocument{%
  \providecommand\BibTeX{{%
    \normalfont B\kern-0.5em{\scshape i\kern-0.25em b}\kern-0.8em\TeX}}}

\setcopyright{acmcopyright}
\copyrightyear{2020}
\acmYear{2020}
\acmDOI{10.1145/doi}

\graphicspath{{figures/artbreeder_images/}}

\begin{document}

\title{Toward Quantifying Ambiguities in Artistic Images}

\author{Xi Wang}
\affiliation{\institution{TU Berlin and MIT}%
\streetaddress{Marchstrasse 23}%
\city{Berlin}%
\postcode{10587}%
\country{Germany}}
\email{xi.wang@tu-berlin.de}

\author{Zoya Bylinskii}
\affiliation{\institution{Adobe Research}%
\streetaddress{1 Broadway} 
\city{Cambridge}
\state{MA} 
\postcode{02142}
\country{USA}}

\author{Aaron Hertzmann}
\affiliation{%
\institution{Adobe Research}%
\streetaddress{601 Townsend St}%
\city{San Francisco}%
\state{CA}%
\postcode{94103}%
\country{USA}}

\author{Robert Pepperell}
\affiliation{%
\institution{Fovolab/Cardiff Metropolitan University}%
\streetaddress{200 Western Avenue}%
\city{Cardiff}%
\postcode{CF5 2YB}%
\country{UK}}


\renewcommand{\shortauthors}{Wang, Bylinskii, Hertzmann, Pepperell}

\begin{abstract}
It has long been hypothesized that perceptual ambiguities play an important role in aesthetic experience: a work with some ambiguity engages a viewer more than one that does not. However, current frameworks for testing this theory are limited by the availability of stimuli and data collection methods. This paper presents an approach to measuring the perceptual ambiguity of a collection of images.  Crowdworkers are asked to describe image content, after different viewing durations. Experiments are performed using images created with Generative Adversarial Networks, using the Artbreeder website. We show that text processing of viewer responses can provide a fine-grained way to measure and describe image ambiguities.
\end{abstract}




\maketitle

\section{Introduction}

\epigraph{\itshape ``When looking at a picture, one should say that the more associations it can open up the better."}{---Pablo Picasso \cite{picassoQuote}}

When confronted with a new image, the human visual system automatically tries to make sense of it 
\cite{potter1975meaning,torralba2009many,SchynsOliva94,oliva2009visual}.
Some images are easy to interpret, but others require more effort because they are ambiguous, multivalent, or indeterminate.  Art theorists have argued that visual art often exploits ambiguity in order to engage viewers by simultaneously suggesting and concealing the meaning of a work, or by evoking multiple diverse meanings \cite{gamboni}, as Picasso suggests in the quote above. 
Time plays an important role in these theories: some images are confusing at first, but then lead to an ``Aha'' moment as the subject is recognized \cite{MuthAha,MuthAha2013}, whereas images that appear simple at first but become more perplexing over time are sometimes called 
``indeterminate" \cite{PepperellVI}.  Indeterminacy is a major theme in modern visual art \cite{gamboni,Pepperell}, e.g., Gerhard Richter is an example of a major contemporary artist who deliberately creates and values indeterminate images \cite{richter} (Fig.~\ref{fig:paintings}).

\newcommand{\ptgheight}{2.25in}

\begin{figure}
\centering
\includegraphics[height=\ptgheight]{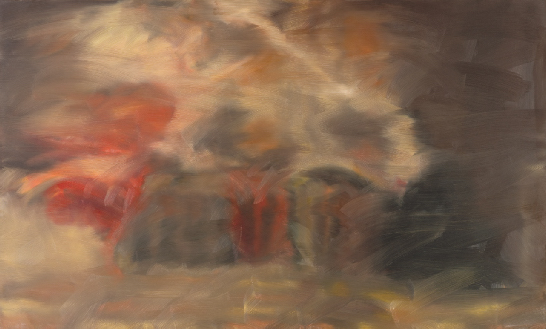}
\hfill
\includegraphics[height=\ptgheight]{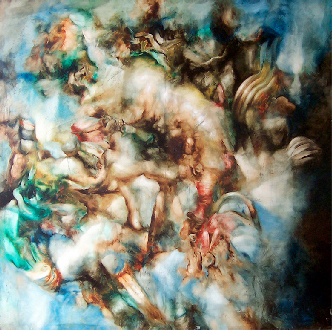}
\caption{Two examples of indeterminate paintings by contemporary artists. 
\textbf{Left}: Annunciation After Titian by Gerhard Richter, Oil on Canvas, 1973 (CR 344-1). © Gerhard Richter 2020 (0099).
\textbf{Right}: Succulus by Robert Pepperell, Oil on Panel, 2005.
\label{fig:paintings}
}
\end{figure}

Can we analyze these image properties quantitatively? Several recent authors have attempted to categorize them \cite{PepperellVI,SeIns,Pepperell,variants}, and to put them in the context of neuroscience theories \cite{VandeCruys,HertzmannScienceOfArt}. 
However, experimentation remains difficult.
Previous studies 
have tended to use  ``handmade'' artworks \cite{SeIns,wallravenBeholder,wallraven_sap,fairhall2008,muth_challenge,fairhall_memory,jakesch}. \newtext{For example, most methods compare artworks made by one artist to works by another artist, or to works by the researchers. Such images} have several limitations when used as experimental stimuli. For example, \newtext{viewers'} judgments may be influenced by historical, stylistic, or contextual factors not of direct relevance to the study. 
\newtext{Highly-simplified graphics, such as Mooney faces, have also been used as stimuli \cite{MuthAha2013}, but generalizing from these results} 
is also challenging. 
Moreover, \newtext{previous} studies typically ask high-level \newtext{subjective} questions of in-person participants, including whether or not an image is ambiguous or contains an object. As a result, the conclusions from these methods are promising but necessarily limited.

This paper proposes an approach to measuring perceptual ambiguity in artistic images. 
Study participants are shown an image for a fixed duration, and asked to describe the contents of the image.  We hypothesized that the quantity and diversity of the descriptions would provide a measure of  perceived ambiguity. 
We also hypothesized that the distribution of responses would vary \newtext{for different viewing durations, reflecting how perception of an image evolves over time}.

For stimuli, we use images from Generative Adversarial Networks (GAN) \cite{GANs}. Specifically, we gather popular images from the website Artbreeder.com, since these images span a range of ambiguity and indeterminacy \cite{HertzmannIndeterminacy}.  These stimuli avoid some of the limitations of previous studies: they are presented in their ``native'' format as digital images, they minimize art historical or contextual confounds, they are visually rich, relatively free of stylistic bias, and can be generated in large numbers.

Understanding ambiguous and indeterminate images is important not only for art theory and history but for our understanding of human image perception more generally. 
We show that histograms and entropy of viewer response can capture and summarize   image ambiguities.
Our results suggest how these kinds of studies could help describe, categorize, and measure the space of image ambiguity. 

\section{Perceptual Study and Data Processing}

\begin{figure*}[t!]
    \centering
    \begin{tabular}{ccc}
        \includegraphics[height=1.3in]{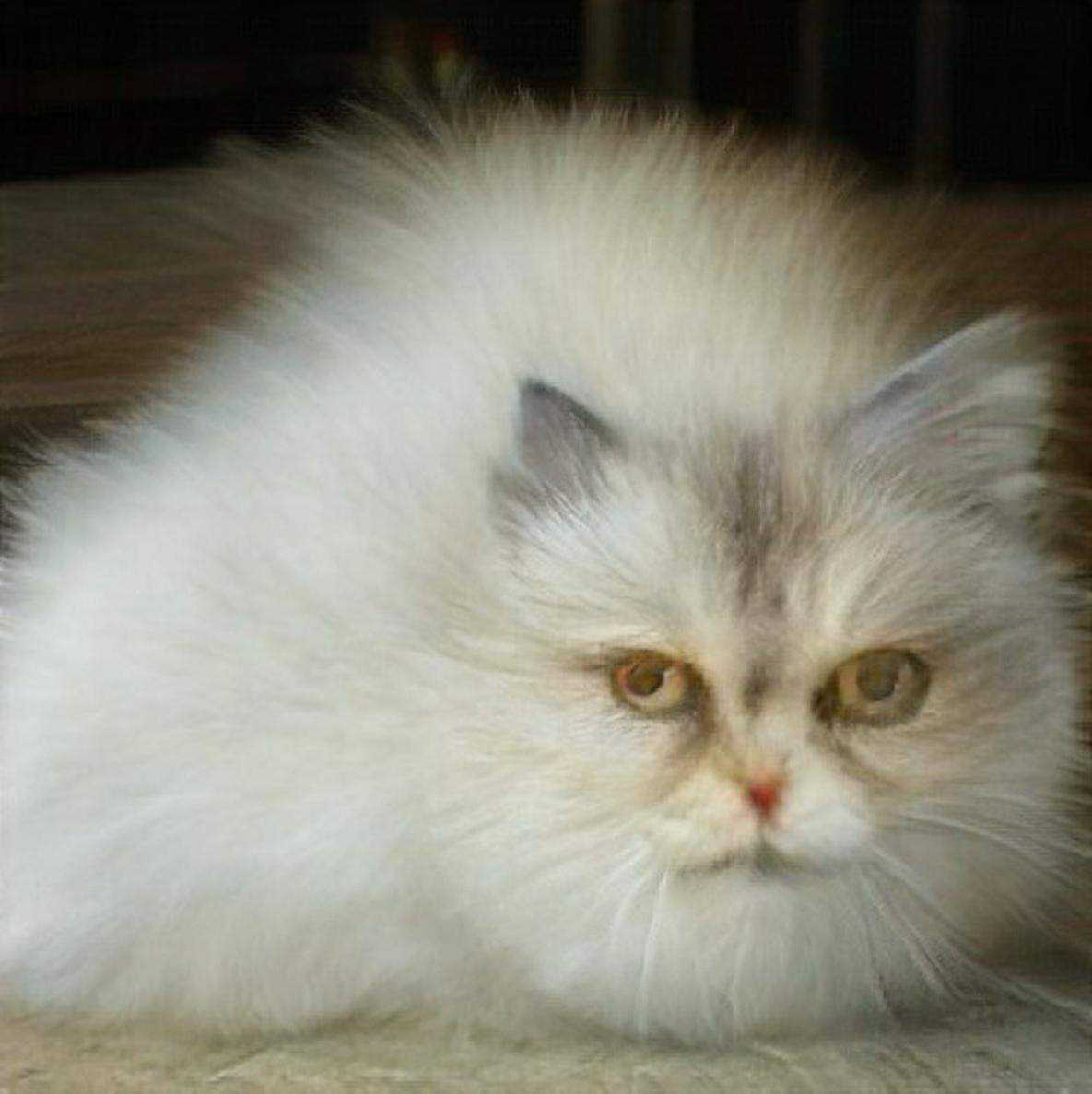}
&
\raisebox{.8in}{
\small
\begin{tabular}{clc}
\textbf{Dur.} & \textbf{Description} & \textbf{Tokens}\\\hline 
0.5s & cat & cat \\
0.5s & cat & cat \\
0.5s & a white cat & cat \\
0.5s & grumpy cat & cat \\
& \vdots\\
3s & white fluffy cat & cat \\
3s & a white cat & cat \\
3s & white  cat & cat \\
3s & cat & cat 
\end{tabular}}
&
    \includegraphics[height=1.45in]{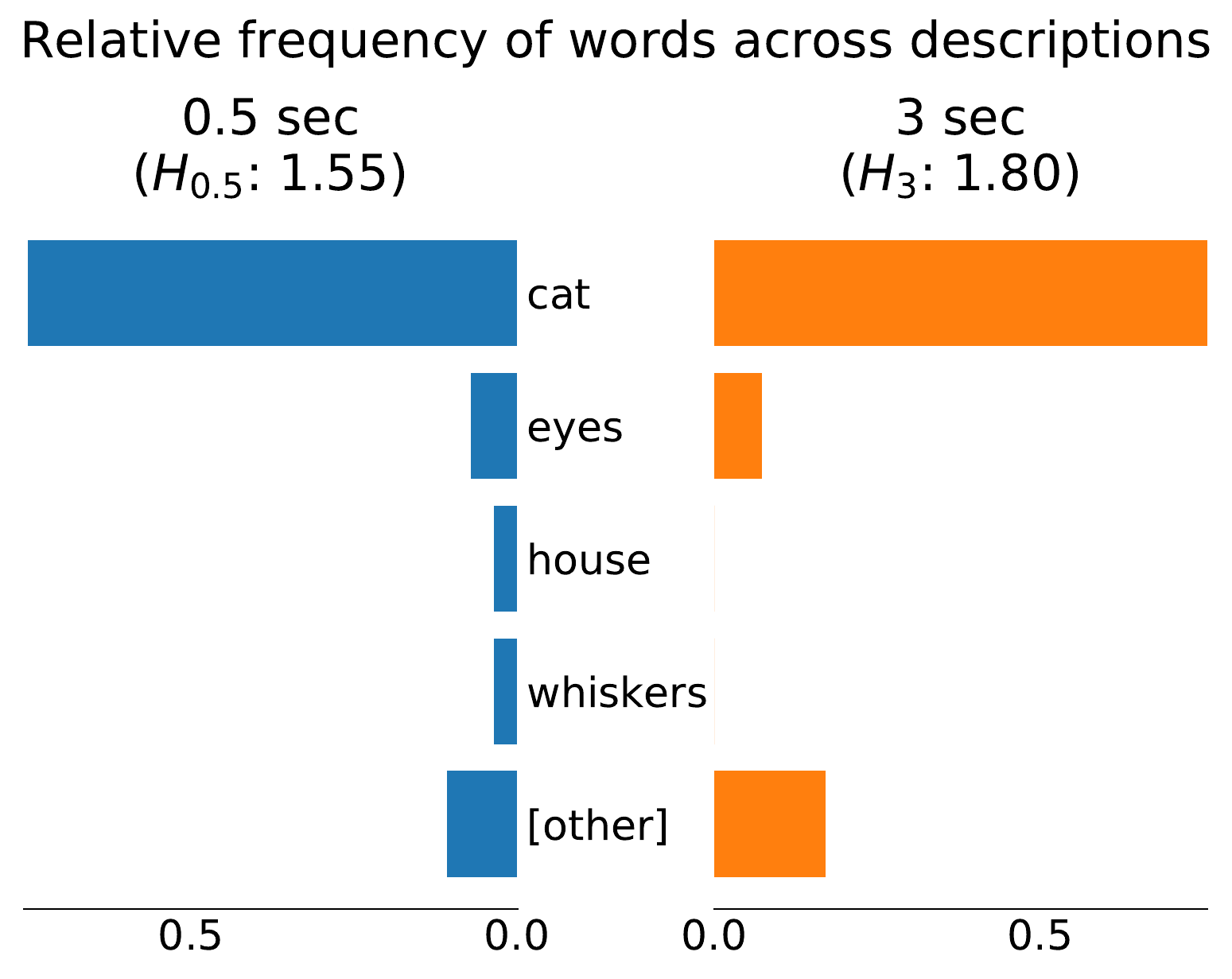} \\
\addlinespace[0.6cm]

    \includegraphics[height=1.3in]{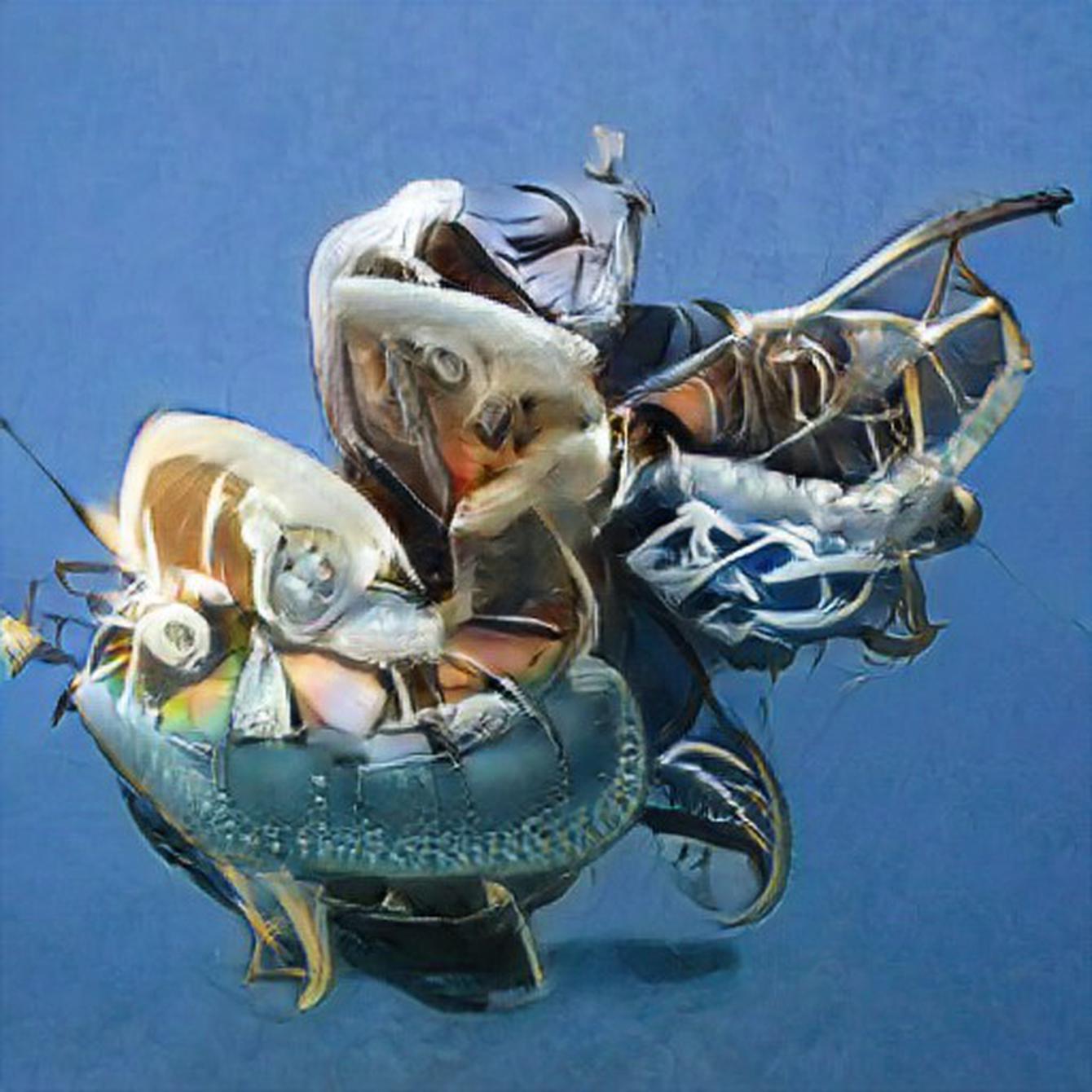}
&
\raisebox{.8in}{
\small
\begin{tabular}{clc}
\textbf{Dur.} & \textbf{Description} & \textbf{Tokens}\\\hline 
0.5s & animal & creature \\
0.5s & shell on legs & legs \\
0.5s & an insect in the sky & insect, sky \\
0.5s & a fish or sea creature & fish, sea, creature \\
& \vdots\\
3s & A calm shell & shell \\
3s & A robot looking swan & robot, swan \\
3s & a giant snail & snail \\
3s & a sea urchin or plant & sea urchin, plant
\end{tabular}}
&
    \includegraphics[height=1.55in]{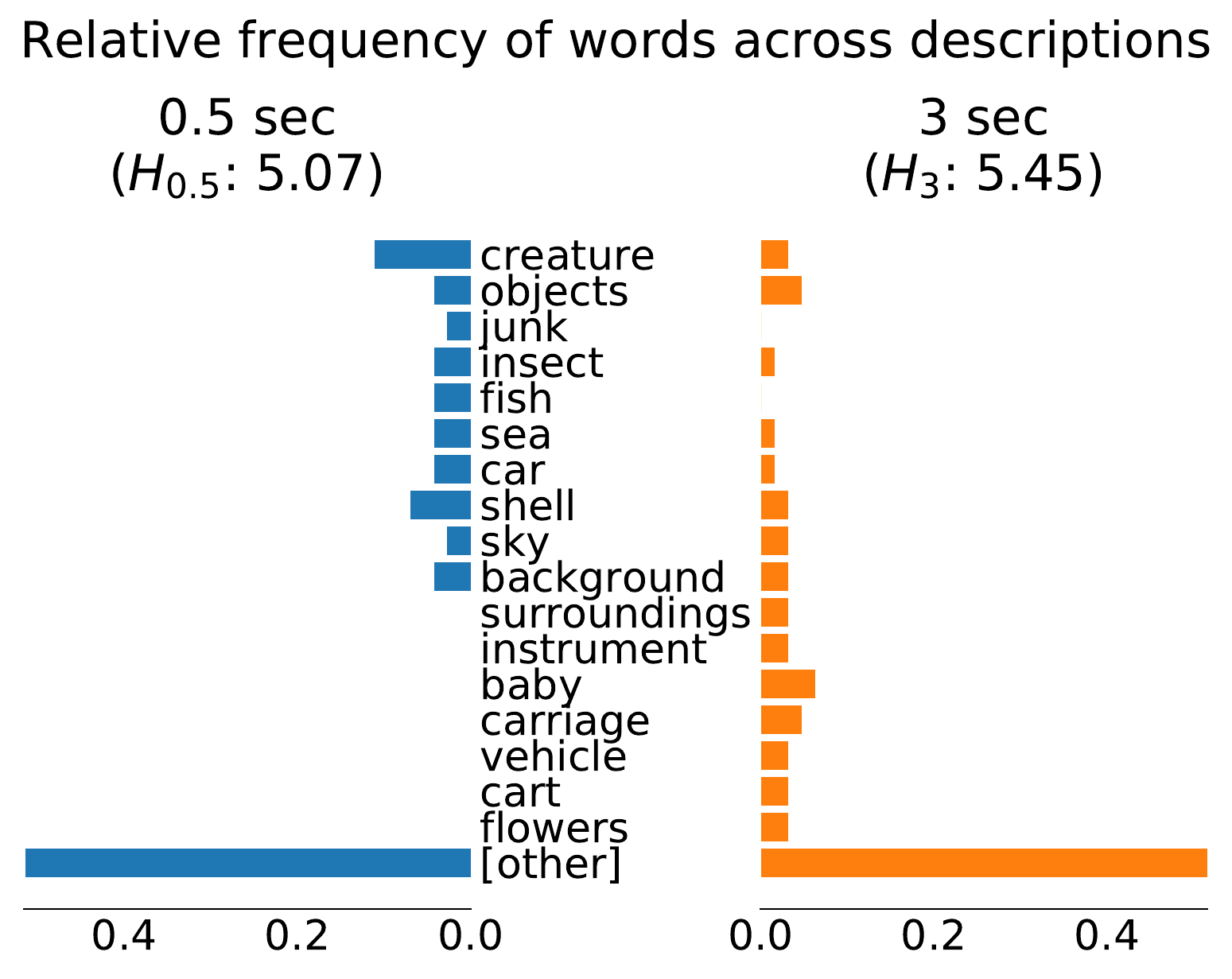} \\
(a) Stimulus image & 
(b) Sampling of image descriptions & 
(c) Description histograms
\end{tabular}
    \caption{Our perceptual study pipeline summarizes the distribution of how participants describe an image after a fixed viewing duration.
    \textit{Top Row:}
    A sample stimulus image is shown in the upper-left. In the middle column, sample descriptions are shown for each image, along with the text tokens extracted by our processing. Note that the descriptions are quite homogenous for the top row: nearly all participants describe the first image as a cat. This is reflected in the histogram of tokens on the upper right, for both time durations.  The entropies are correspondingly low: $H_{0.5}=1.55$, $H_{3}=1.80$.
    The "[other]" histogram bin counts tokens that  appeared only once each.
    \textit{Bottom row:}
    A more indeterminate image yields much more variability in descriptions, and the variability increases considerably over time: $H_{0.5}=5.07$, $H_3=5.45$.
    }
    \label{fig:pipeline}
\end{figure*}

\emph{Image stimuli:} To form the image dataset, we manually identified a set of 150 images from Artbreeder.com that appeared to exhibit variations in image ambiguity.  All images were taken from Artbreeder's ``General'' class, which provides images from the BigGAN model \cite{biggan}. The first 120 images were manually selected from among the most popular images on Artbreeder, and loosely categorized into 4 categories of 30 images each: ``Recognizable'' (e.g., Fig.~\ref{fig:pipeline}(top)), ``Dichotomous'' (\newtext{depicting two or more distinct objects simultaneously, e.g.,} Fig.~\ref{fig:morefigures}(d,e)), ``Indeterminate'' (\newtext{open to multiple interpretations, e.g.,} Fig.~\ref{fig:pipeline}(bottom)), and "Abstract'' \newtext{(clearly containing no objects, e.g., Fig.~\ref{fig:morefigures}(a,f))}. We manually constructed the remaining thirty, ``AbstractFlat'', (e.g., Figure \ref{fig:morefigures}(g,h)) to be highly abstract, using the site's ``gene editing'' interface, 
\newtext{by increasing the BigGAN truncation parameter via the ``chaos gene'' control and setting all presented embedding coordinates to -1.}


\emph{Task design:} We crowdsourced descriptions using Amazon's Mechanical Turk. A task consisted of a sequence of 30 images, all from a single category of the stimuli set, to avoid confounding effects. Participants viewed each image for either 0.5 or 3 seconds; the viewing duration was fixed for a single participant, but randomized across participants. After the image disappeared, participants completed an attention vigilance task whereby they reported the last location on the screen where they looked (based on a similar task design~\cite{Fosco_2020_CVPR}). Then, participants entered a freeform text description. They were instructed to describe the scene and any objects in the image, ``even if the image looks abstract at first''. We recruited 70 participants for each of the two viewing durations and 5 categories from our stimuli set (launching 700 tasks in total). We filtered participants based on the attention vigilance task. After filtering, we have on average 20.4 
descriptions \newtext{per} image \newtext{and viewing duration}. Sample images and descriptions are shown in Figures \ref{fig:pipeline}(a,b)\newtext{, and \ref{fig:entropy_ranking}}.

\emph{Viewing duration:} Human perception studies have shown that retention of visual details plateaus by 3 seconds of perception \cite{brady2008visual,brady2013visual}, while half a second allows for most visual information to enter conceptual working memory without interference \cite{potter1999understanding,feifeiJOV}. We ran a pilot study with viewing durations of 300 ms, 500 ms, 1000 ms, and 3000 ms. Participants complained that 300 ms was too brief to understand what was depicted, while results at 1000 and 3000 ms were very similar. Thus, for our main experiment, we chose to collect data at two durations: 500 ms and 3000 ms. 



\emph{Post-processing:} Given the raw textual descriptions of a given image, we perform some simple text processing to form a histogram of responses (Figure \ref{fig:pipeline}(c)).  We treat the set of responses as a ``bag-of-words" \cite{plsi}. 
Specifically, for each text description, we run a part-of-speech tagger \cite{boro-dumitrescu-burtica:2018:K18-2}, and keep only the nouns. We also discard any terms in a predefined set of 12 disallowed words, such as ``abstract" and ``art''. 
Synonyms are grouped using NLTK \cite{nltk}, yielding a set of tokens.  
We then form a histogram of the tokens for the image, grouped by viewing duration. This process is performed separately for each image's responses. 

Given these histograms, we measure ambiguity with two numbers: $H_{0.5}$ is the Shannon entropy of the 0.5-second viewing duration histogram, and $H_{3}$ is the entropy of the 3-second histogram. We report entropy scores in units of bits.


\section{Categorizing and ranking ambiguities}

Our key assumption is that the distribution of textual descriptions for a given image and a given time duration reflect the perceptual ambiguity that a single viewer has for that image. 
This could arise, for example, if, when forced to describe an image, a viewer samples a single possible explanation from their posterior probability distribution. Similar processes have been hypothesized elsewhere in neuroscience \cite{dawPigeon}. 

Hence, the histogram of responses acts as an estimate of a typical viewer's probability distribution over image interpretations, for a given viewing duration.  We can observe several types of images. In a determinate image, (Figure \ref{fig:pipeline}(top row)), most viewers describe the image in the same way at both viewing durations; that is, both $H_{0.5}$ and $H_3$ are low. In an indeterminate image (Figure \ref{fig:pipeline}(bottom row)), descriptions are highly varied in both conditions. 

Figure \ref{fig:entropy_ranking} shows the images with the lowest and highest entropy \newtext{across descriptions}. As can be seen in the figure, \newtext{this metric} reflects the degree of recognizability of the images.  \newtext{Moreover, note that none of the ``AbstractFlat'' images appear here. Sometimes they have high entropy (e.g., Figure \ref{fig:morefigures}(g)), and sometimes they have lower entropy due to a peak of color descriptions like ``pink image'' (e.g., Figure \ref{fig:morefigures}(h)). This suggests that perhaps entropy alone can identify indeterminate images. An image which is too abstract does not conjure many associations, nor does an image which is very realistic. Indeterminacy seems to produce the longest and most varied descriptions.}

Figure \ref{fig:difference_ranking} shows high-entropy images sorted by the entropy difference $H_3 - H_{0.5}$ under the condition of $H_3>4$. This threshold  gives us images which generate more associations after 3s viewing. When sorted by the entropy difference, we see \newtext{that more complex images tend to have greater decrease in entropy over time. The complementary examples are shown in Figure \ref{fig:difference_ranking_comp}, where the threshold is set to $H_3<4$. Here the images all seem to depict individual objects. Entropy appears to decrease the most where the object is odd but recognizable. Entropy appears to increase for images that seem to be complex variations on familiar objects.}


 \newcommand{\rfht}{1.1in}
 \newcommand{\rfsep}{\hspace{1.0mm}}
 \begin{figure*}
 \centering
  \begin{tabular}{p{\rfht}p{\rfht}p{\rfht}p{\rfht}p{\rfht}}
 \includegraphics[width=\rfht]{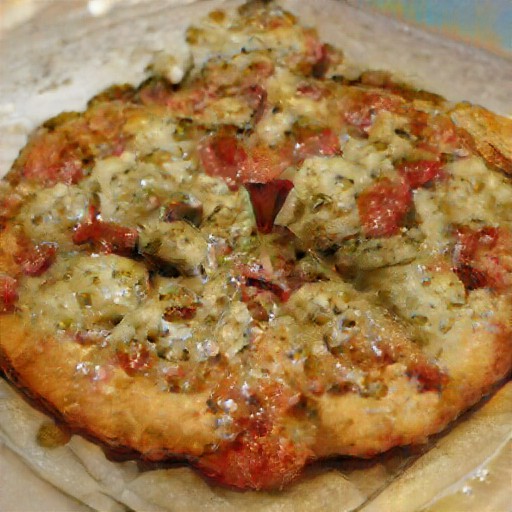} &
\includegraphics[width=\rfht]{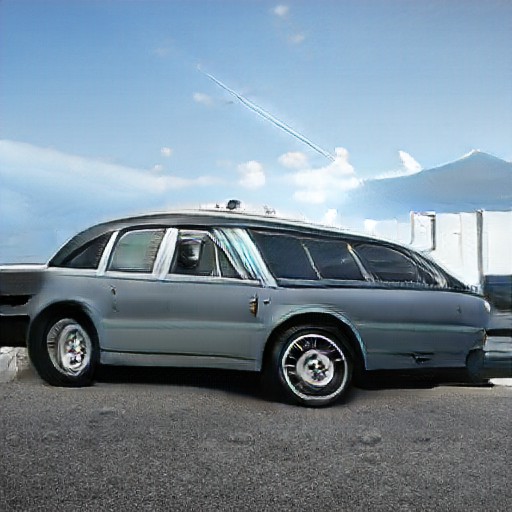} &
\includegraphics[width=\rfht]{fe10e169420cb497ec3719f8.jpeg} &
\includegraphics[width=\rfht]{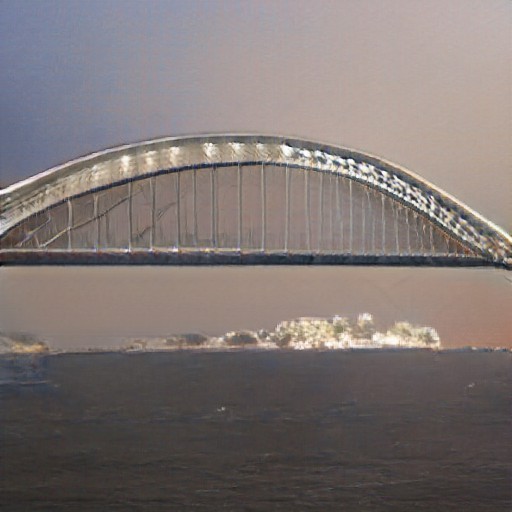} &
\includegraphics[width=\rfht]{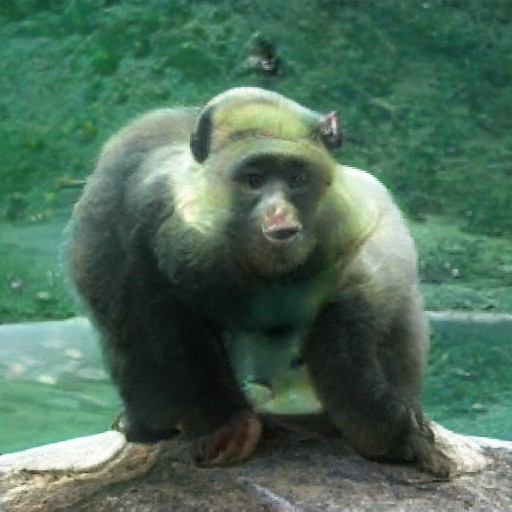} \\
\small{$H_3$}: 1.05 &
\small{$H_3$}: 1.76 &
\small{$H_3$}: 1.8 &
\small{$H_3$}: 2.00 &
\small{$H_3$}: 2.10 \\
\small{pizza, weird pizza, small pizza, small pizza with toppings} & \small{car, blue car, distorted car, a weird blue limo looking car} & \small{white fluffy cat, a white cat, white cat, cat} & \small{bridge, bridge over water, a bridge and a sea, a bridge} & \small{monkey, a gorilla, an overweight monkey, deformed monkey} \\[2ex]
\includegraphics[width=\rfht]{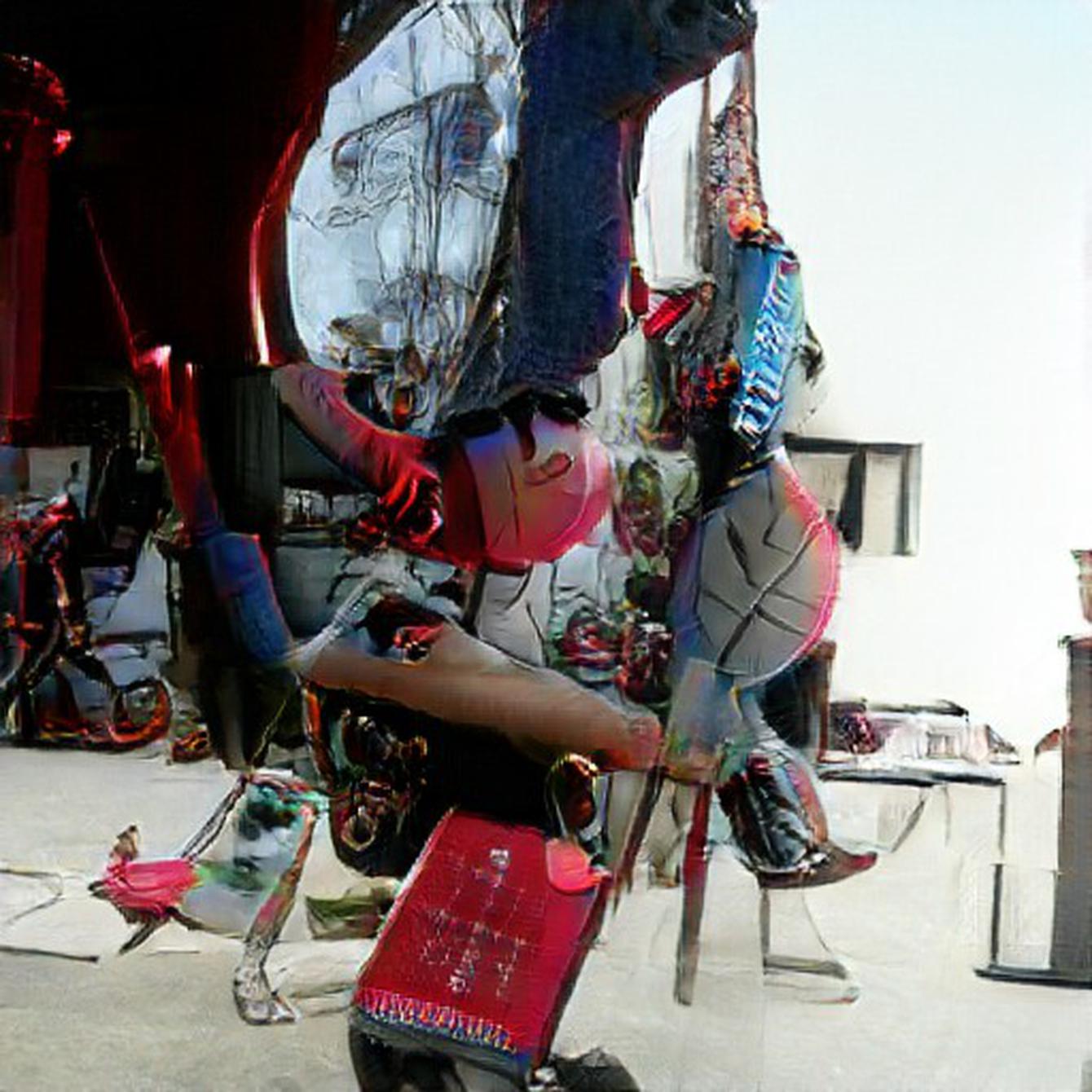} &
\includegraphics[width=\rfht]{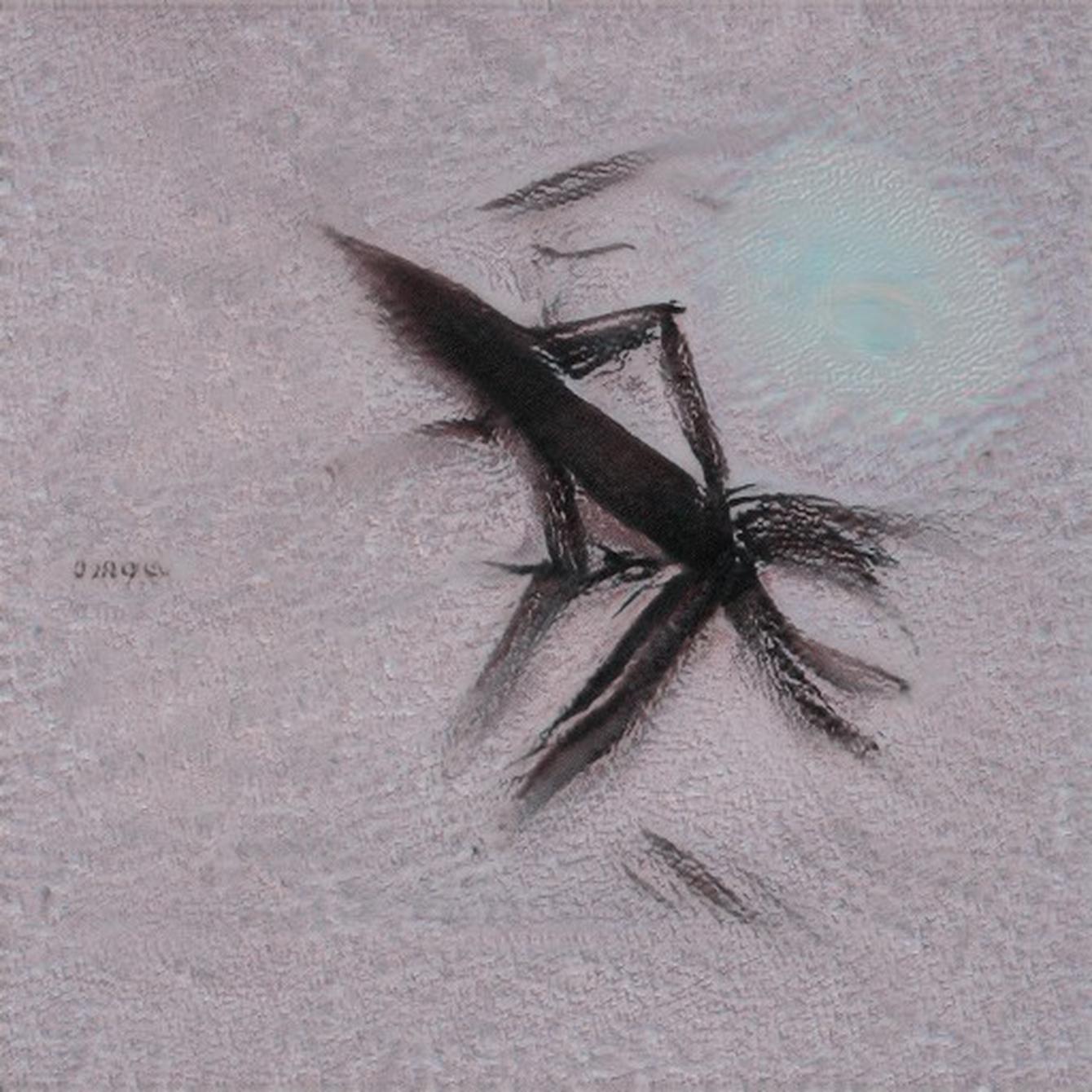} &
\includegraphics[width=\rfht]{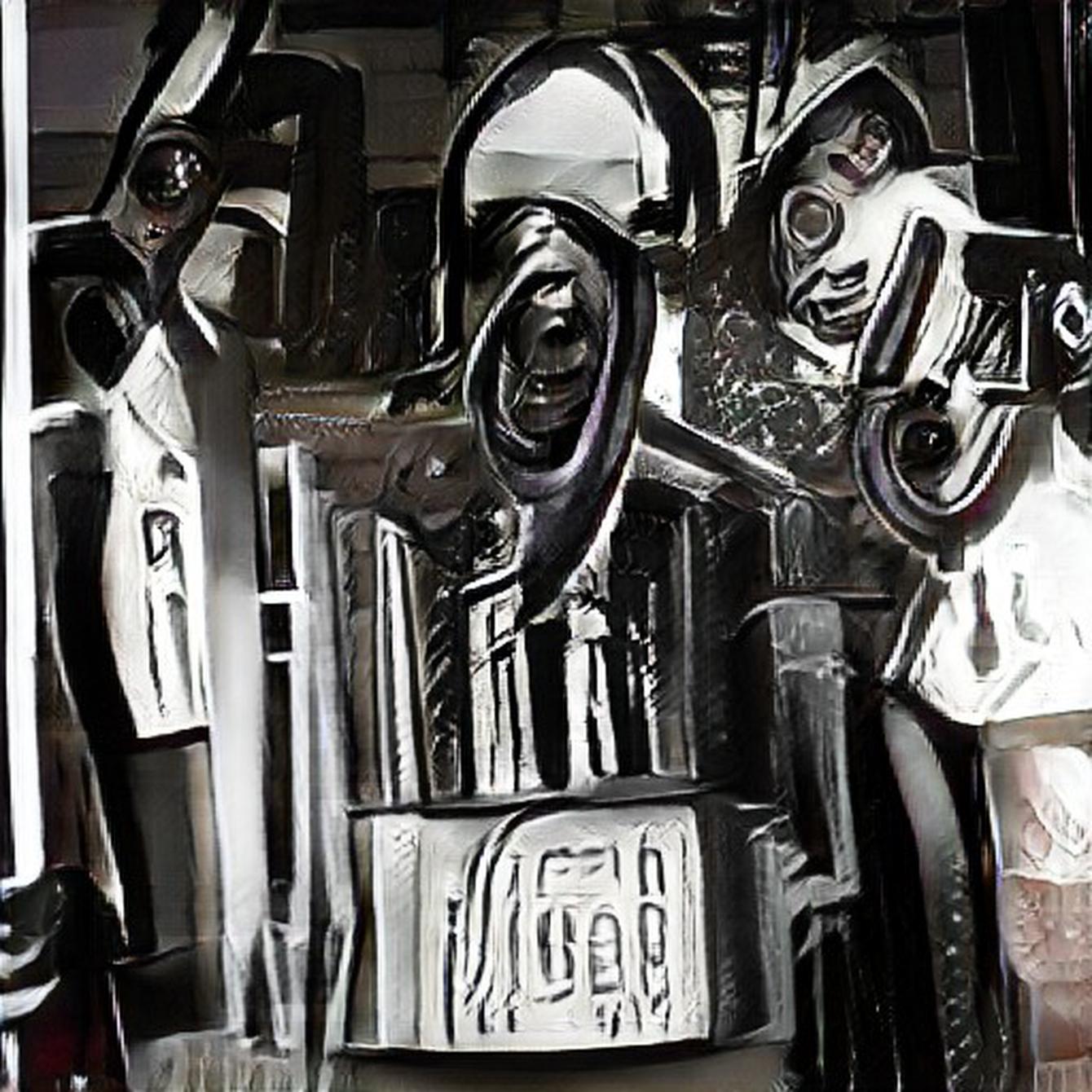} &
\includegraphics[width=\rfht]{231aa61a9390b25dab4c.jpg} &
\includegraphics[width=\rfht]{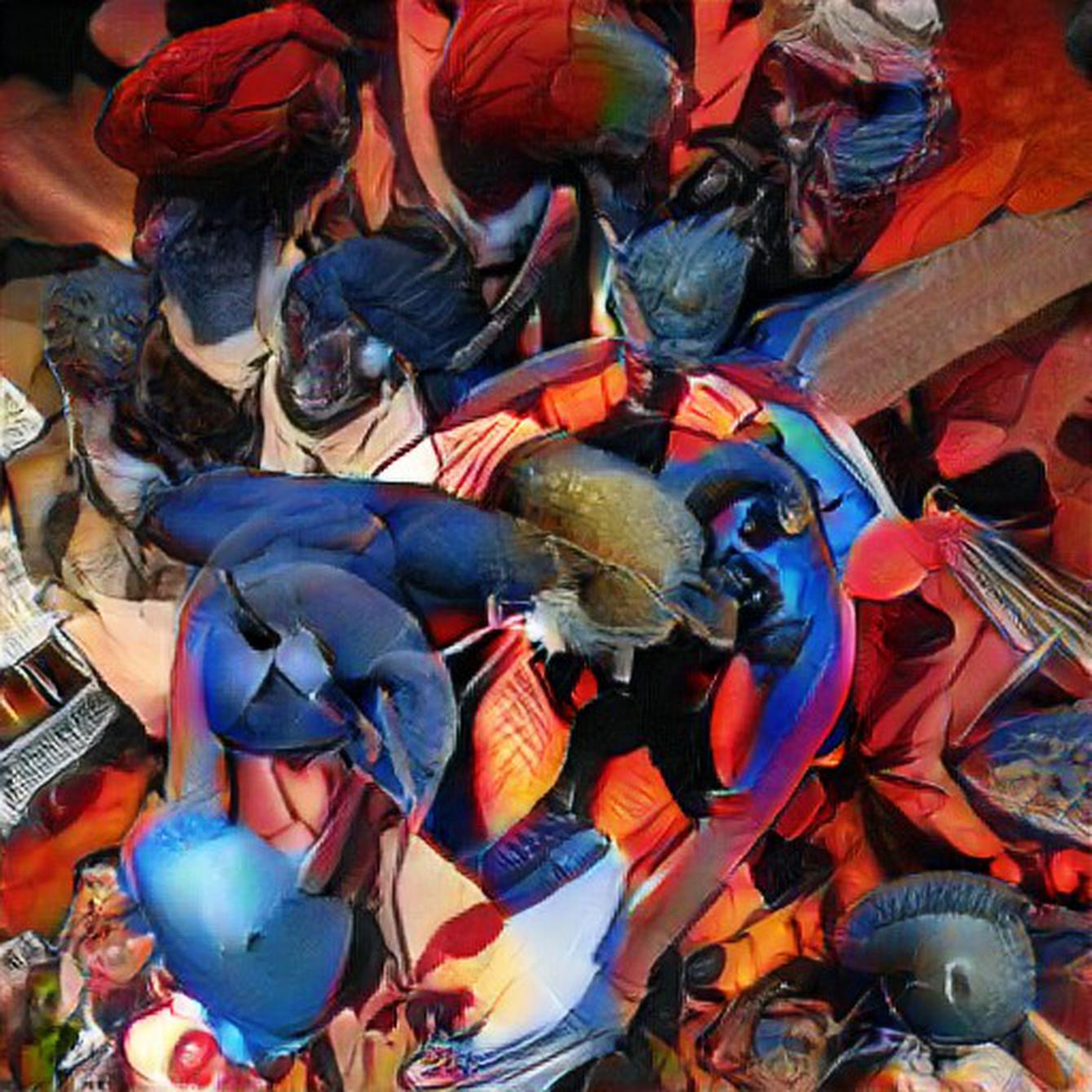} \\
\small{$H_3$}: 5.37 &
\small{$H_3$}: 5.38 &
\small{$H_3$}: 5.43 &
\small{$H_3$}: 5.45 &
\small{$H_3$}: 5.70 \\
\small{mechanics garage, a mangled person, statue of technology, parade costume} & \small{fishing bait hook, clippings from a haircut, chinese writing, black feathers} & \small{steel work, metal suit of armor, three tin men hanging out, a radiator} & \small{a robot looking swan, musical instrument, a sea urchin or plant, a weird looking seashell} & \small{flower buds, stained glass, some sort of bunched up cloth, autumn leaves}
 \end{tabular}
 \caption{The images in our dataset with the lowest and highest entropy ($H_3$) \newtext{of descriptions. Some sample descriptions are included.  Observe that entropy appears to reflect image ambiguity/recognizability, and very indeterminate images have the highest entropy.}  }

 \label{fig:entropy_ranking}
 \end{figure*}

\newcommand{\drfht}{1.1in}
\newcommand{\drfsep}{\hspace{1.0mm}}
\begin{figure*}
\centering
\begin{tabular}{c@{\drfsep}c@{\drfsep}c@{\drfsep}c@{\rfsep}c}
\includegraphics[width=\drfht]{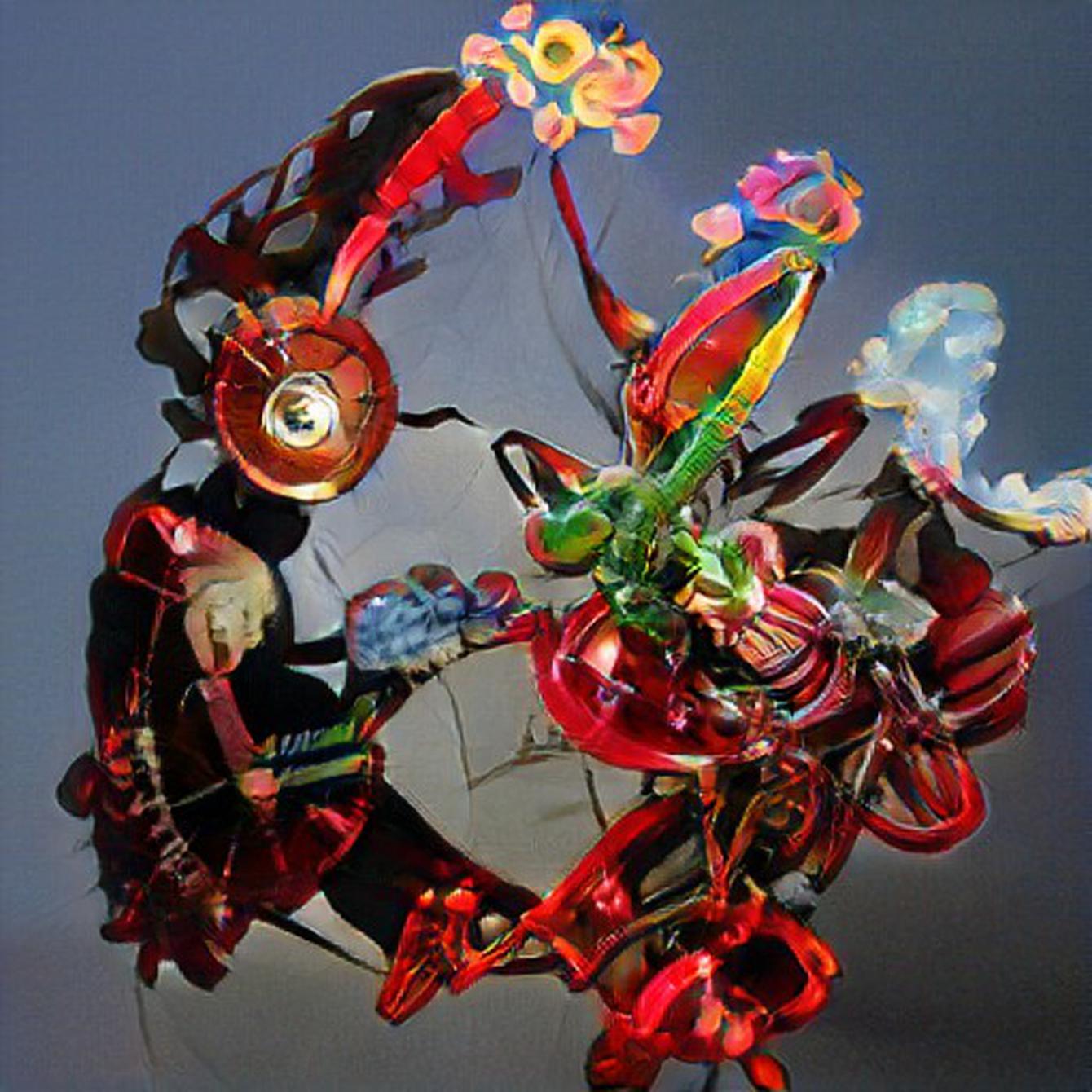} &
\includegraphics[width=\drfht]{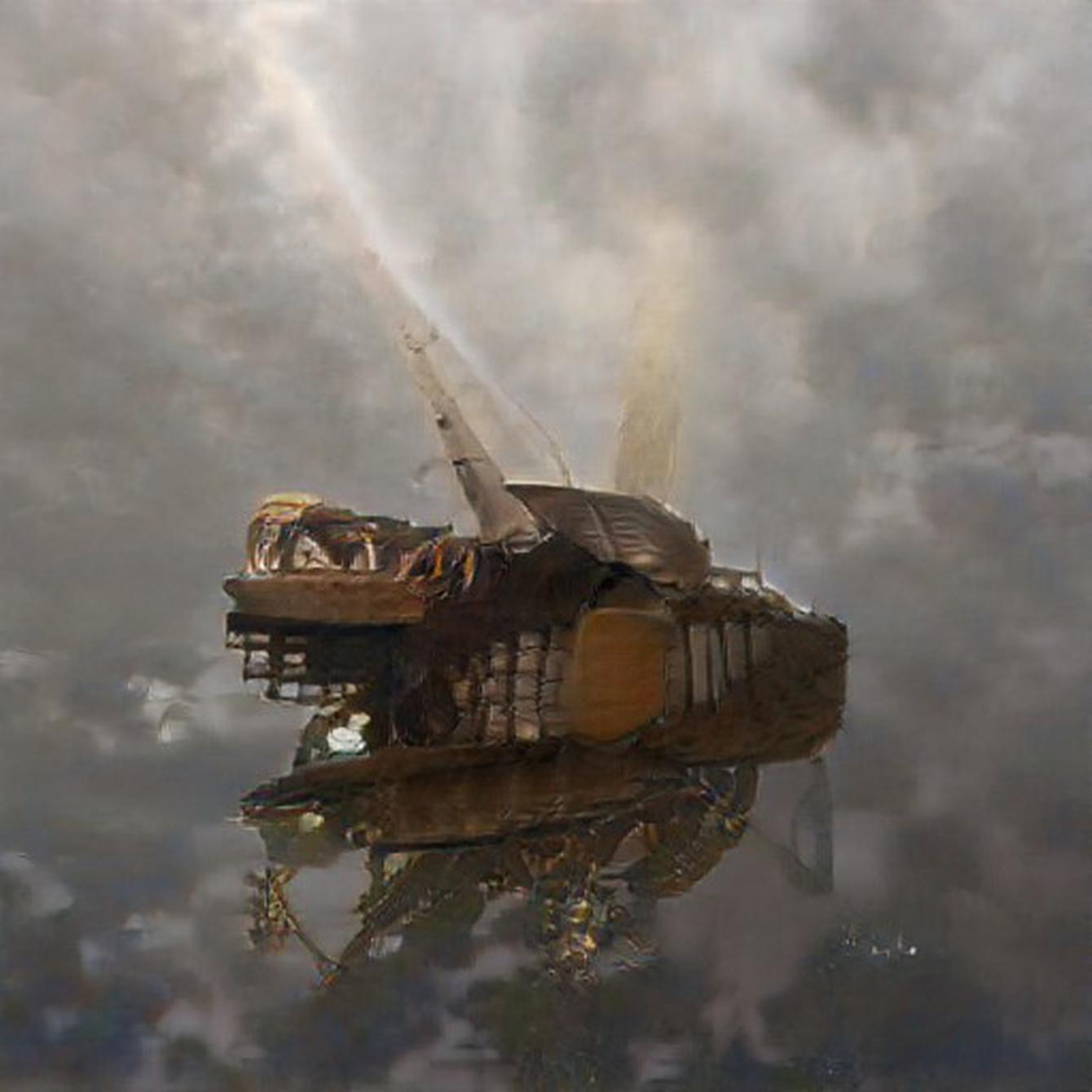} &
\includegraphics[width=\drfht]{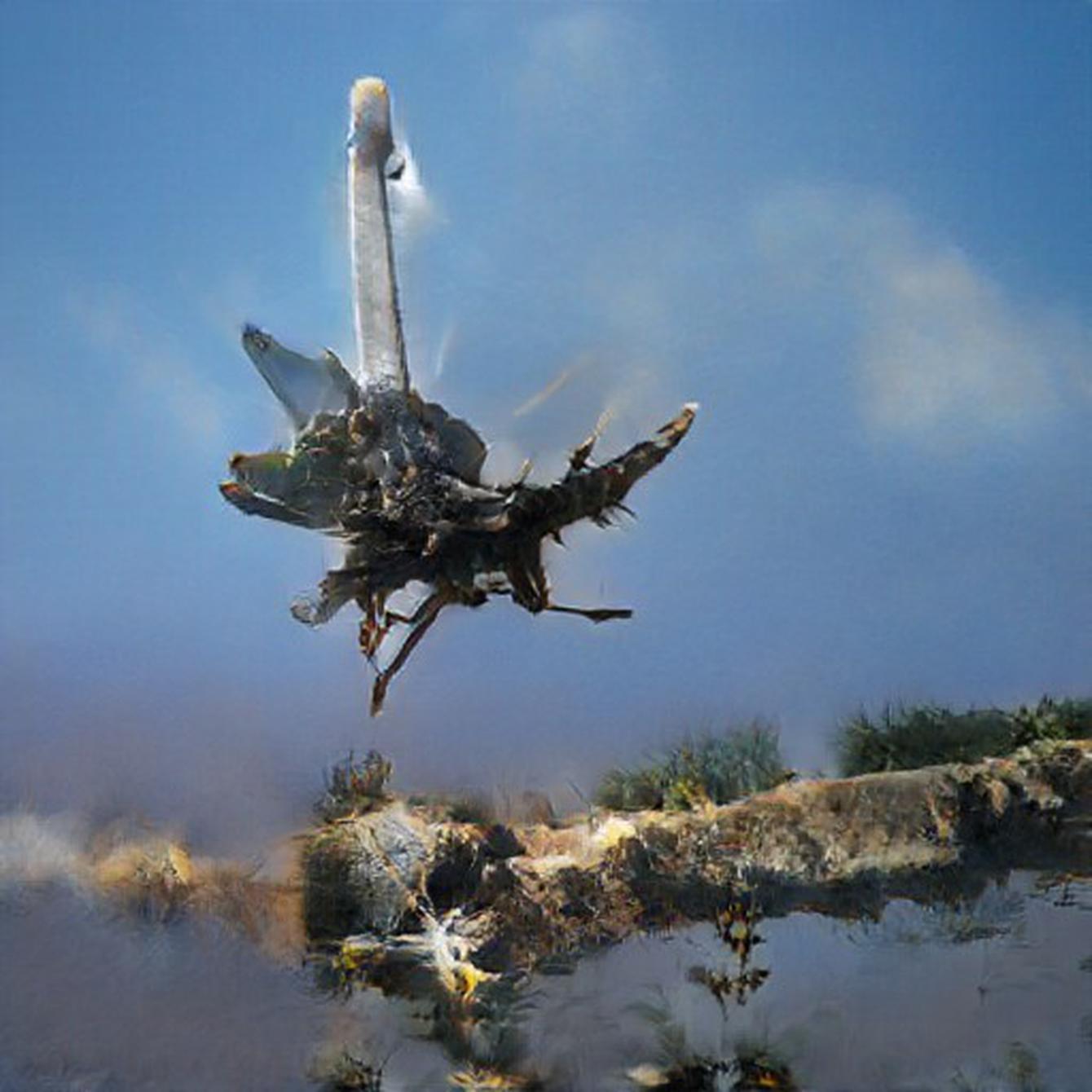} &
\includegraphics[width=\drfht]{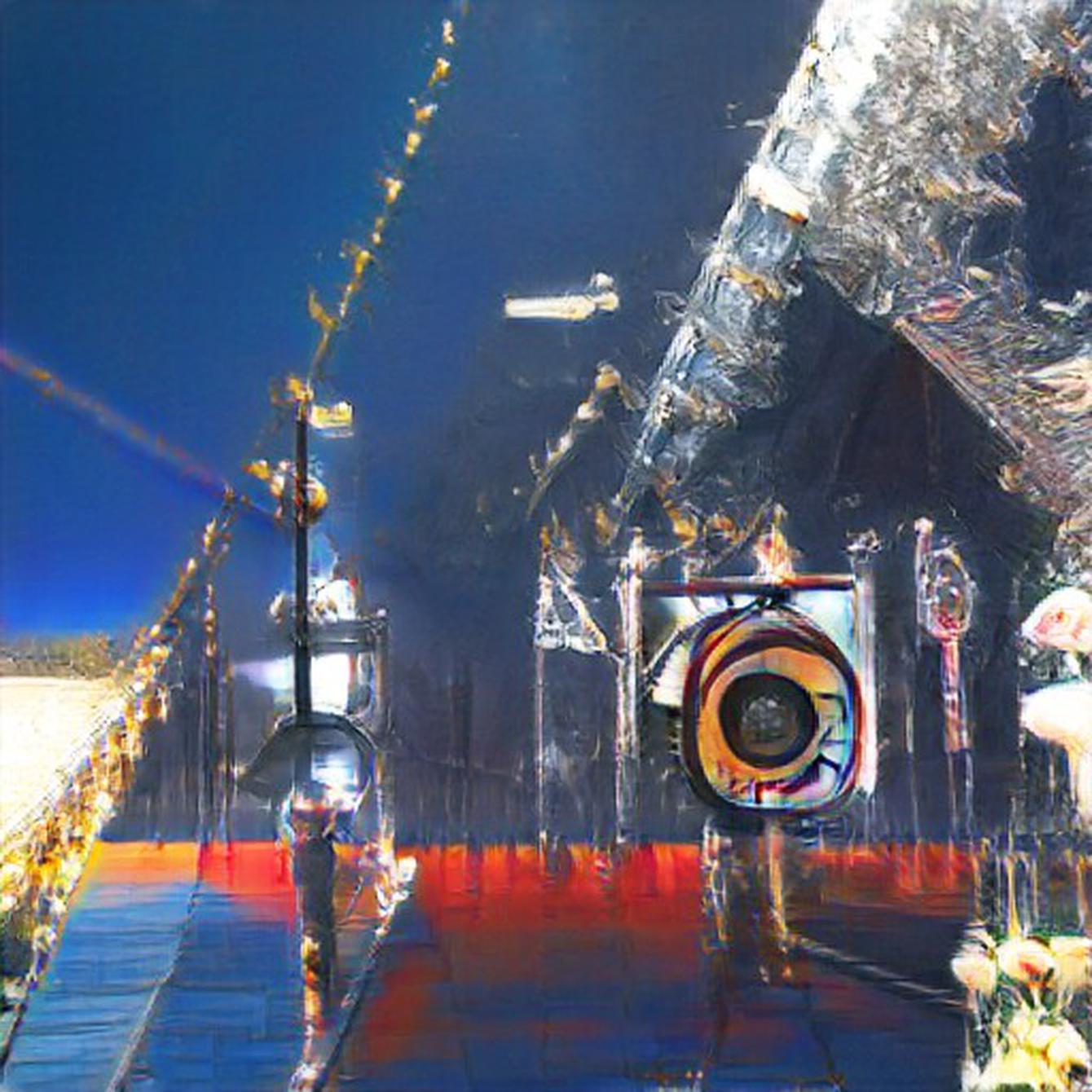} &
\includegraphics[width=\drfht]{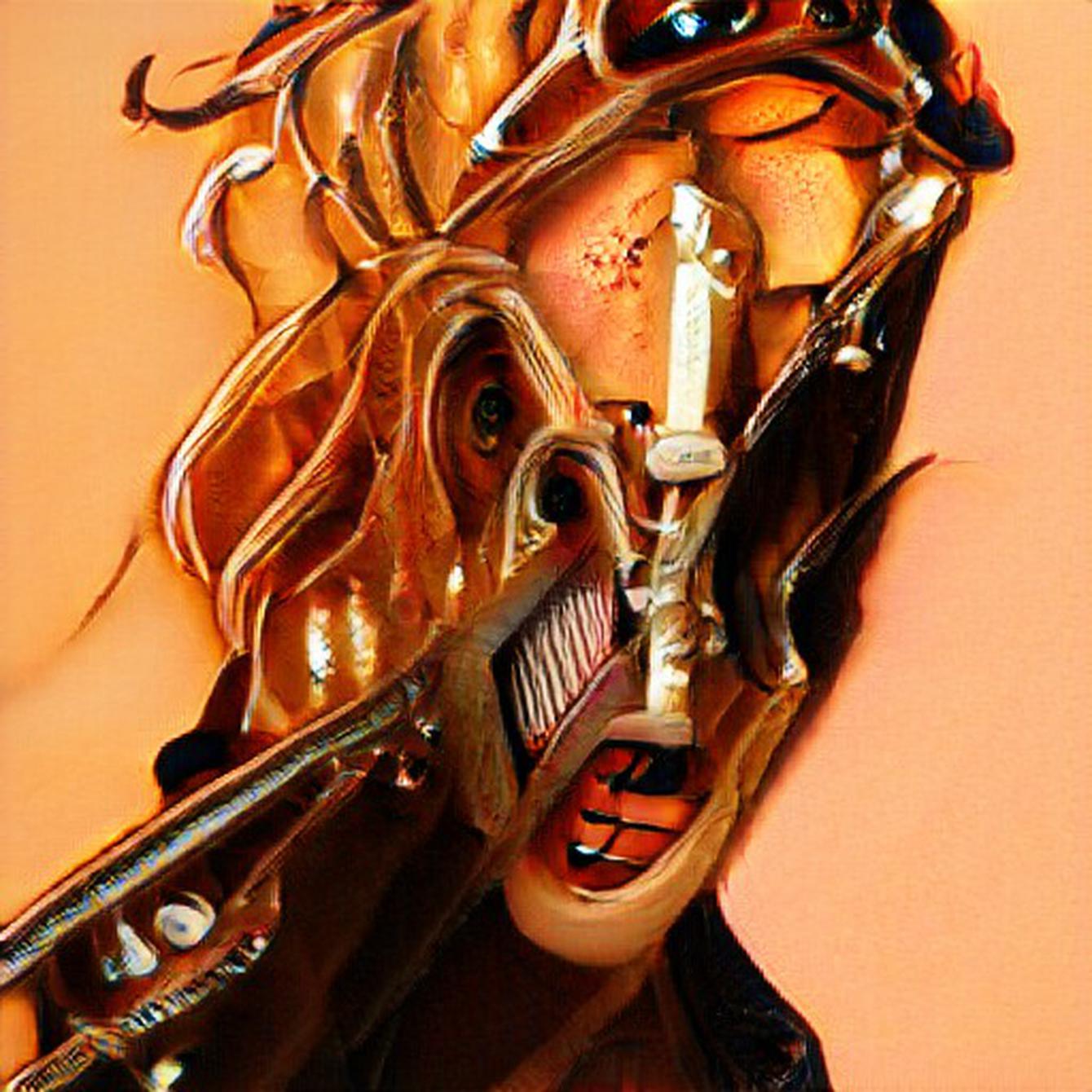} \\
\small{$\Delta H$}: -0.60 &
\small{$\Delta H$}: -0.59 &
\small{$\Delta H$}: -0.59 &
\small{$\Delta H$}: -0.51 &
\small{$\Delta H$}: -0.51 \\
[2ex]
%
%
\includegraphics[width=\drfht]{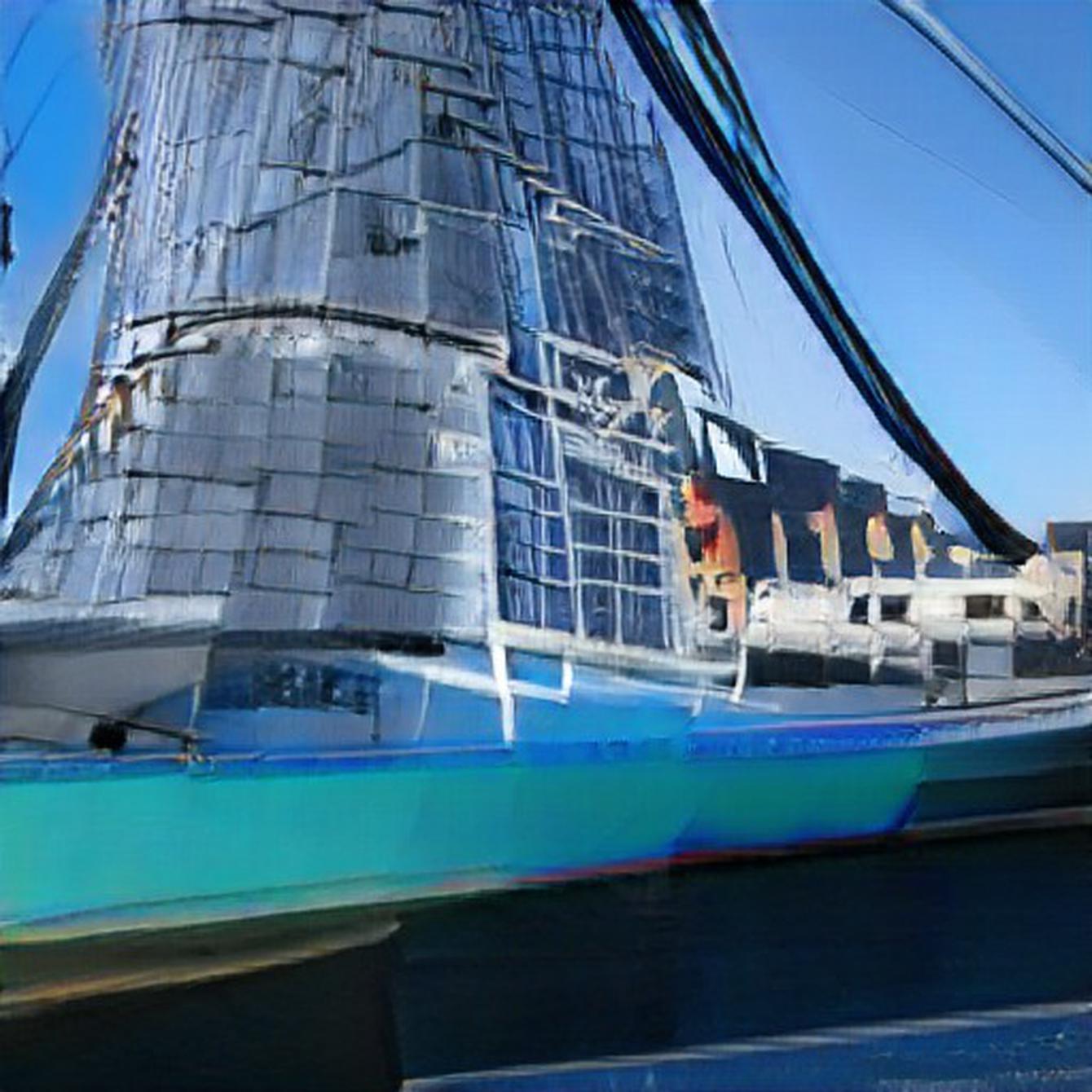} &
\includegraphics[width=\drfht]{0b3e4f1b1e6ad0bd520a.jpeg} &
\includegraphics[width=\drfht]{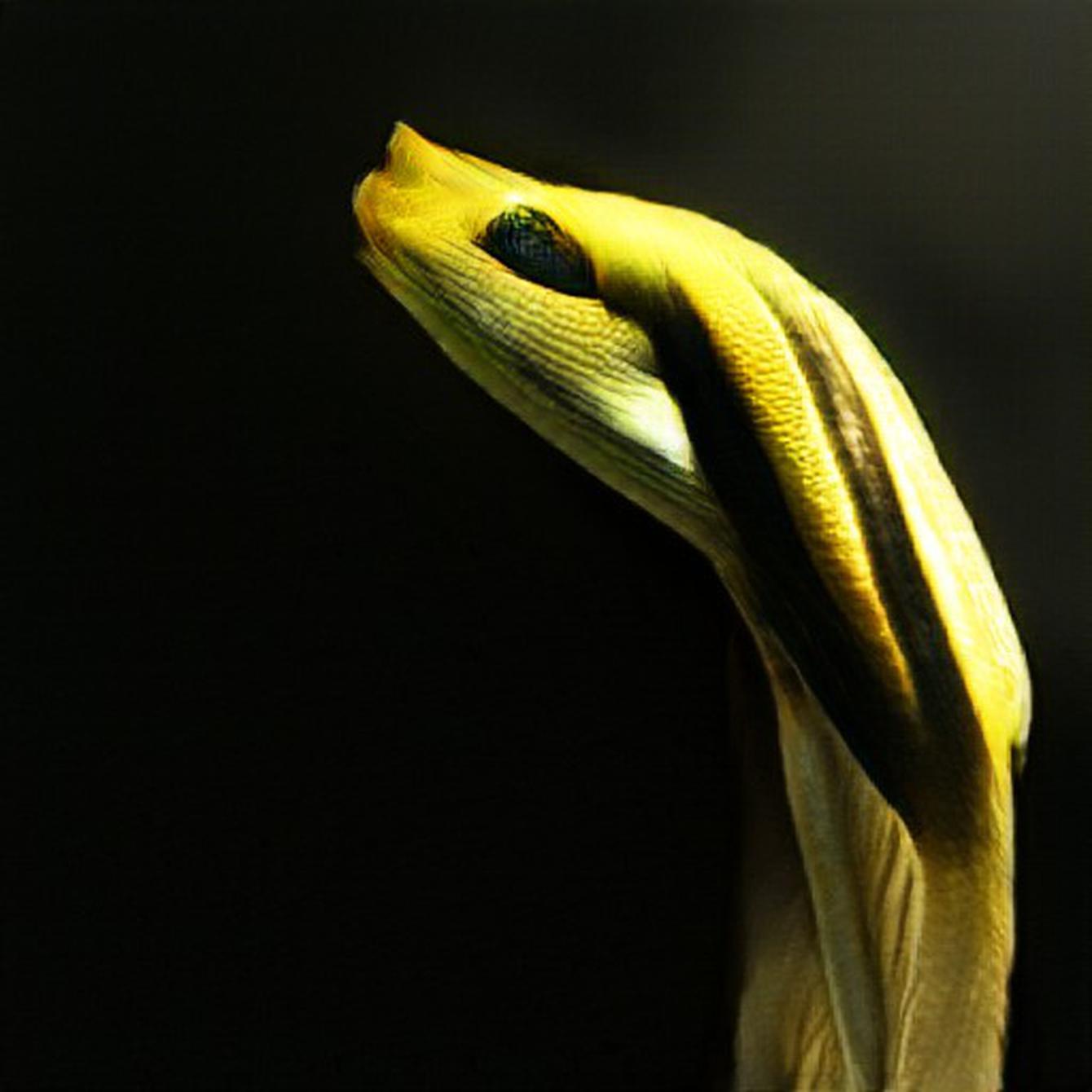} &
\includegraphics[width=\drfht]{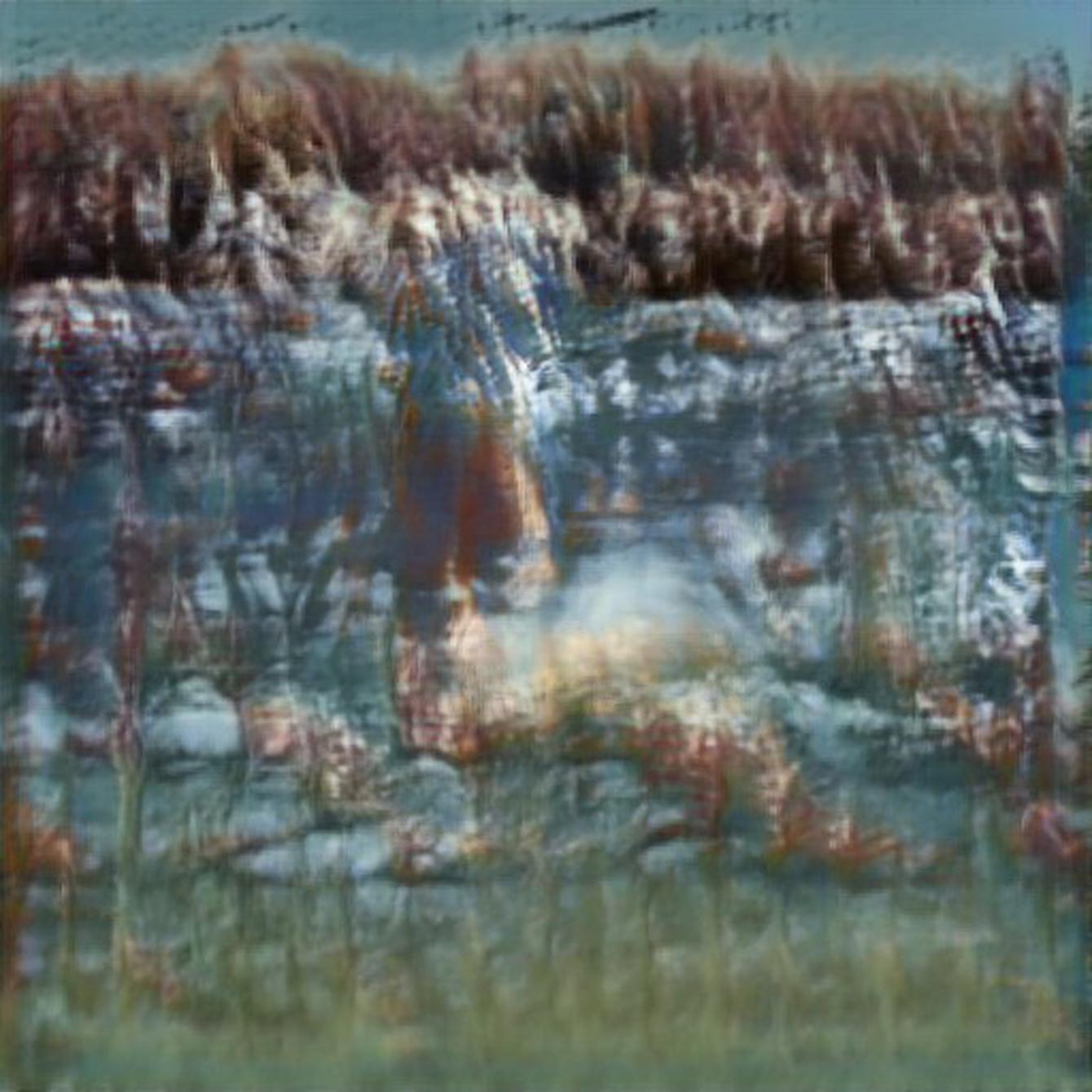} &
\includegraphics[width=\drfht]{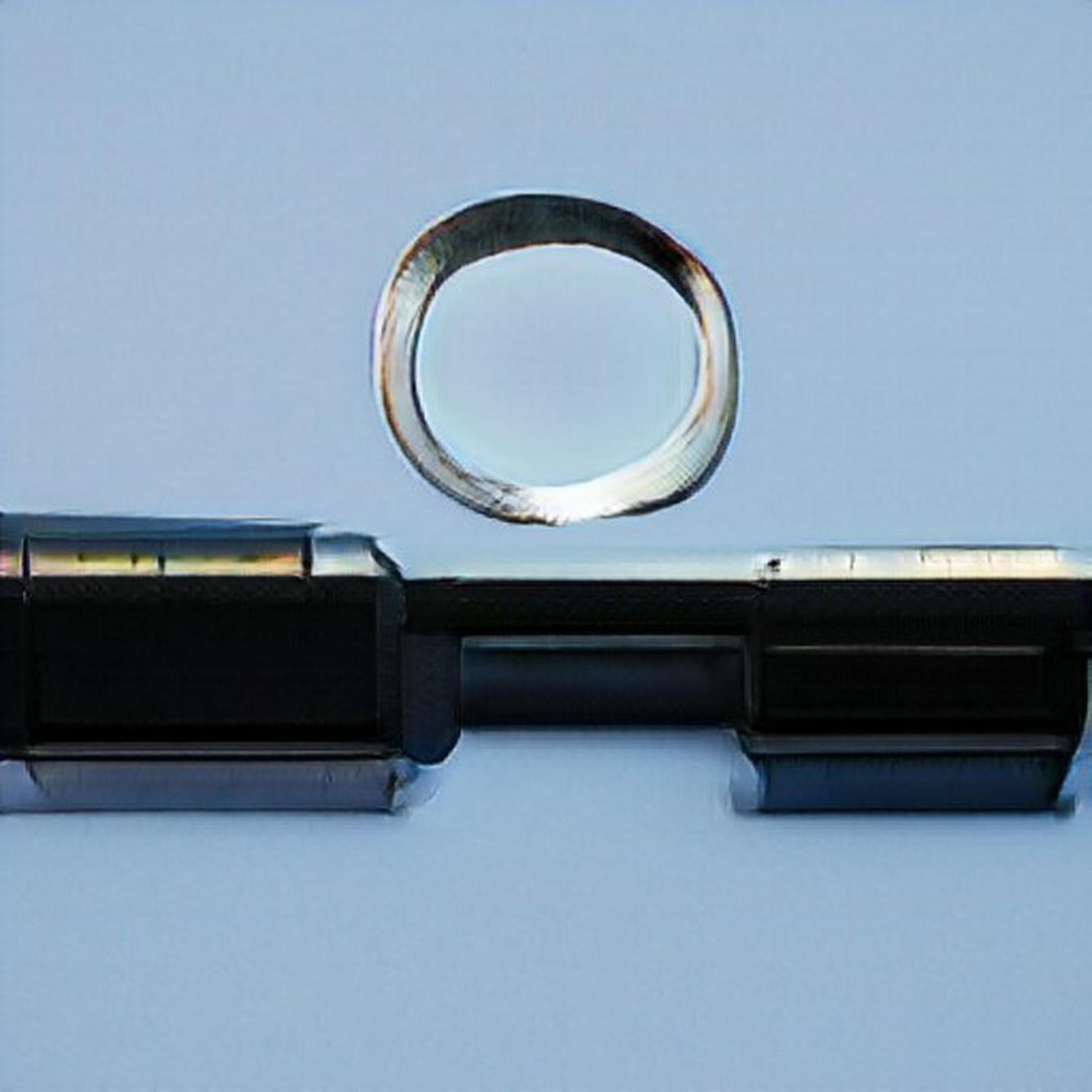} \\
\small{$\Delta H$}: 1.08 &
\small{$\Delta H$}: 1.16 &
\small{$\Delta H$}: 1.16 &
\small{$\Delta H$}: 1.31 &
\small{$\Delta H$}: 1.83 \\
\end{tabular}
\caption{Of the images in our dataset with high \newtext{description} entropy ($H_3 > 4.0$),
the images with the lowest and highest change in entropy ($\Delta H = H_3-H_{0.5}$). \newtext{For the top row images, entropy decreased over time, and, for the bottom row, increased over time.}
\vspace{1ex}
}
\label{fig:difference_ranking}
\end{figure*}

\begin{figure*}
\centering
\begin{tabular}{c@{\drfsep}c@{\drfsep}c@{\drfsep}c@{\rfsep}c}
\includegraphics[width=\drfht]{73a1343f382e17cbe8eb.jpeg} &
\includegraphics[width=\drfht]{93ad84e0988d1674e0f2.jpg} &
\includegraphics[width=\drfht]{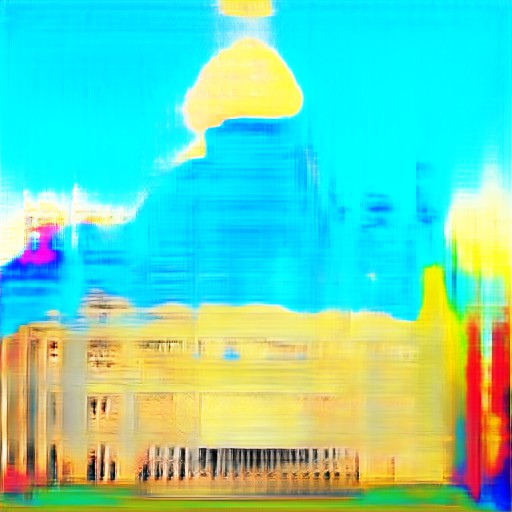} &
\includegraphics[width=\drfht]{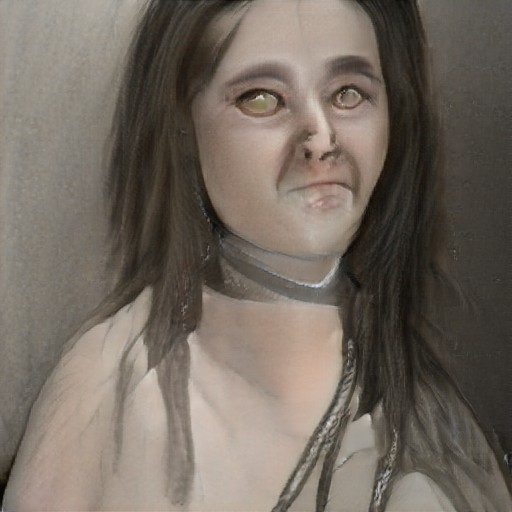} &
\includegraphics[width=\drfht]{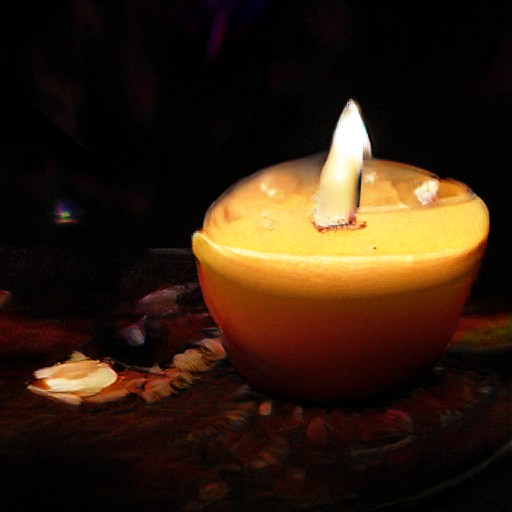} \\
\small{$\Delta H$}: -1.27 &
\small{$\Delta H$}: -0.99 &
\small{$\Delta H$}: -0.79 &
\small{$\Delta H$}: -0.73 &
\small{$\Delta H$}: -0.71 \\[2ex]

\includegraphics[width=\drfht]{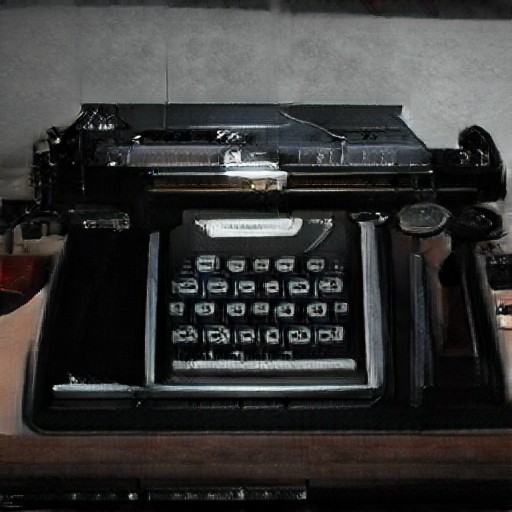} &
\includegraphics[width=\drfht]{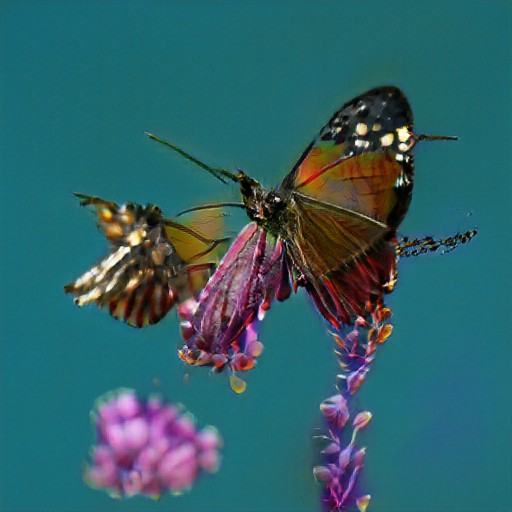} &
\includegraphics[width=\drfht]{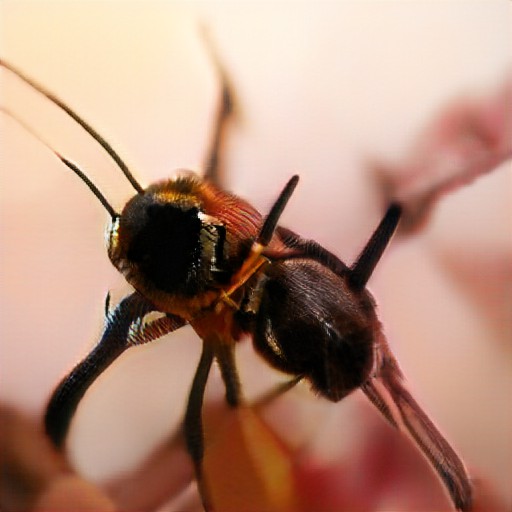} &
\includegraphics[width=\drfht]{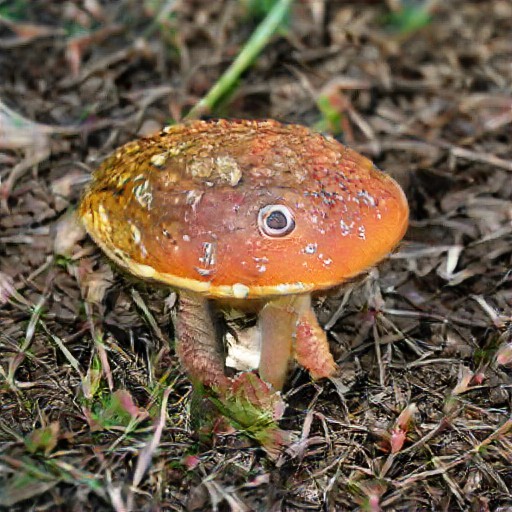} &
\includegraphics[width=\drfht]{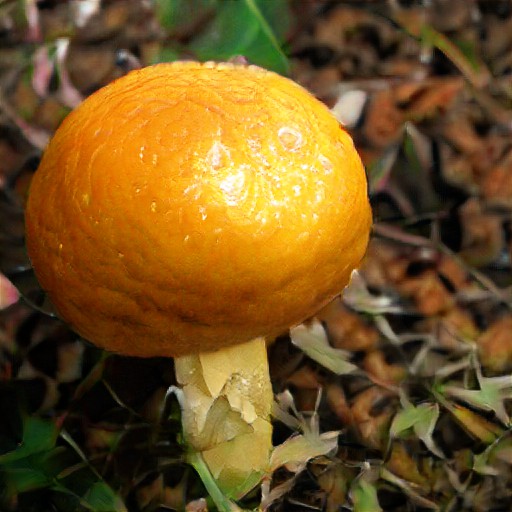} \\
\small{$\Delta H$}: 0.55 &
\small{$\Delta H$}: 0.83 &
\small{$\Delta H$}: 0.86 &
\small{$\Delta H$}: 0.94 &
\small{$\Delta H$}: 1.38 
%
%
\end{tabular}
\caption{\newtext{Of the images in our dataset with low or medium description entropy ($H_3 < 4.0$),
the images with the lowest and highest change in entropy ($\Delta H = H_3 - H_{0.5}$). Note that images with low $\Delta H$ tend to be recognizable, eventually, while large $\Delta H$ images tend to be variations on familiar concepts.} 
}
\label{fig:difference_ranking_comp}
\end{figure*}

\begin{figure*}
    \centering
    \includegraphics[width=4in]{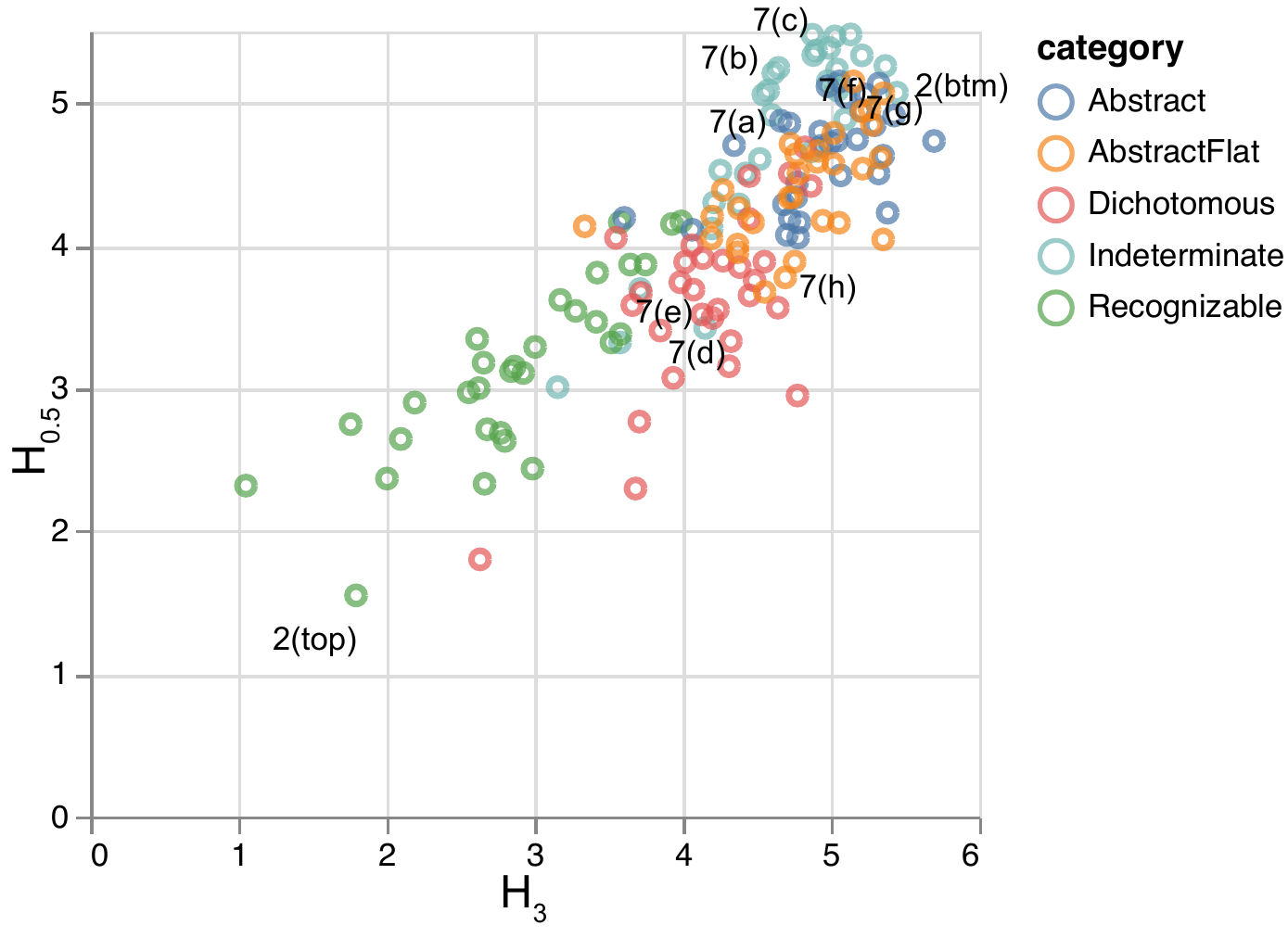}
    \caption{Scatterplot of entropy values in our dataset. The five colors show the loose categorization used to build the dataset. Observe that the entropy values cluster some categories well, including recognizable ($H_{0.5},H_3 < 4$) and dichotomous ($H_{0.5} < 4, H_3>3.5$); the indeterminate images mostly  cluster toward the top.  The locations of images from other figures in the paper are annotated in the plot.
  }
    \label{fig:scatterplot}
\end{figure*}

Figure \ref{fig:scatterplot} shows a scatterplot of the entropies of images across our dataset.  As shown in the plot, the image categorization that we used to build the dataset emerges in some regions of the plot; for example, recognizable images generally have lower entropy scores ($H_{0.5}, H_3 < 4)$. Dichotomous images, such as Figure \ref{fig:morefigures}(d,e) have simple explanations at first, but diverge as viewers find two or more interpretations.

Figure \ref{fig:morefigures} shows a sampling of cases with high entropy. We now discuss some of the phenomena that emerge.  The first three images have high entropy ($H_{3} > 4.5$), and entropy decreases over time ($H_{0.5}>H_3$). In Figure \ref{fig:morefigures}(a), 
the most-frequent terms in the 0.5s viewing condition are ``coat'', ``knife'', and ``cloth'', but, in the 3s condition, entirely new terms become prevalent, including ``building'', ``ship'', and ``sails''. 
In Figure \ref{fig:morefigures}(b)
most descriptions in the 0.5s condition include ``face'', \newtext{and even more do in the 3s condition, while several terms, like ``bug'', ``helmet'', and ``monster'', disappear}. In Figure \ref{fig:morefigures}(c), ``flowers'' is rare in the initial condition, but becomes much more common after the longer viewing duration.

\newcommand{\hwfigwidthtwo}{1.2in}
\newcommand{\hwhistwidthtwo}{1.8in}

\begin{figure*}
\begin{tabular}{cc}
\includegraphics[width=\hwfigwidthtwo]{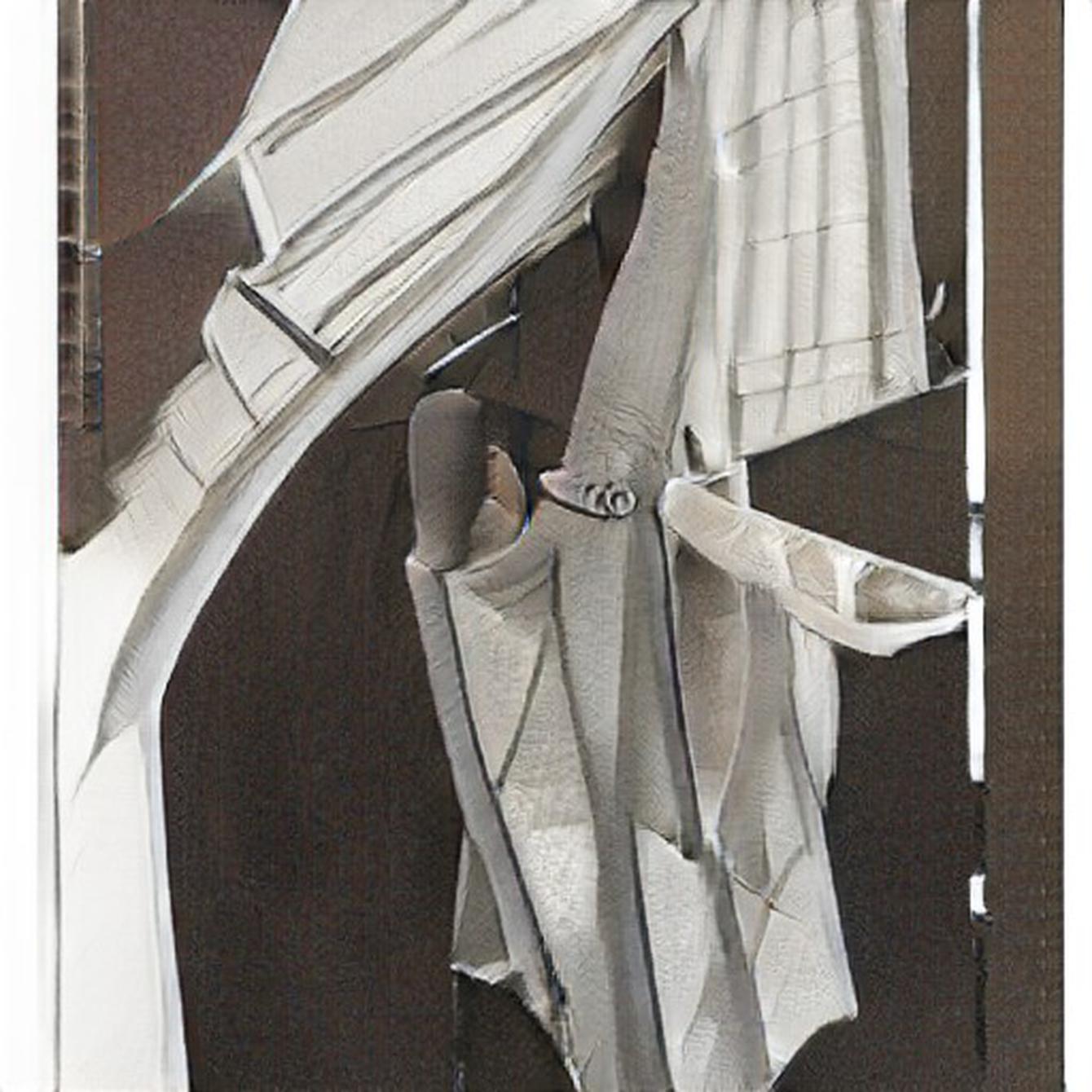}
\includegraphics[width=\hwhistwidthtwo]{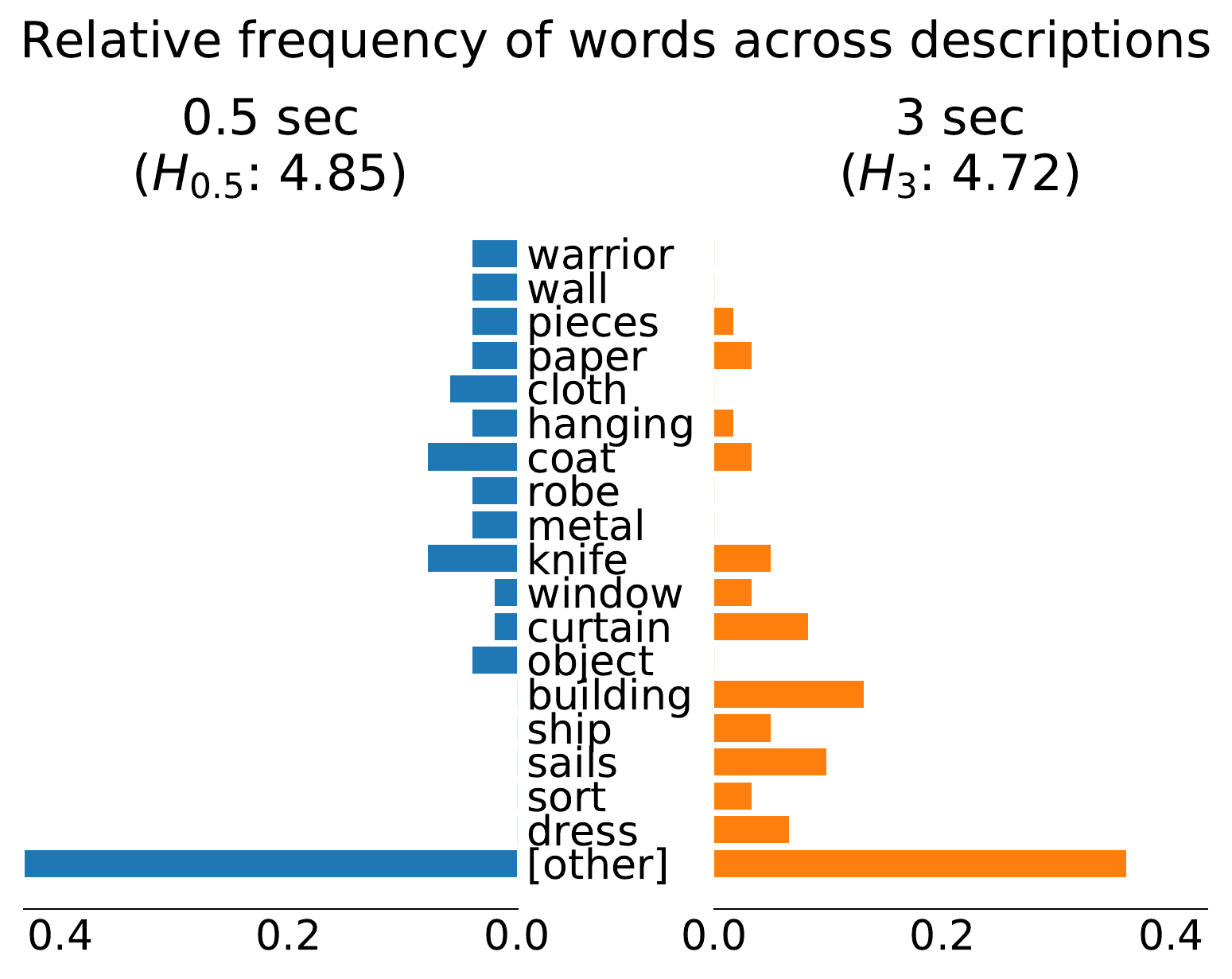}

&

\includegraphics[width=\hwfigwidthtwo]{4dd597e195862a88165c.jpg}
\includegraphics[width=\hwhistwidthtwo]{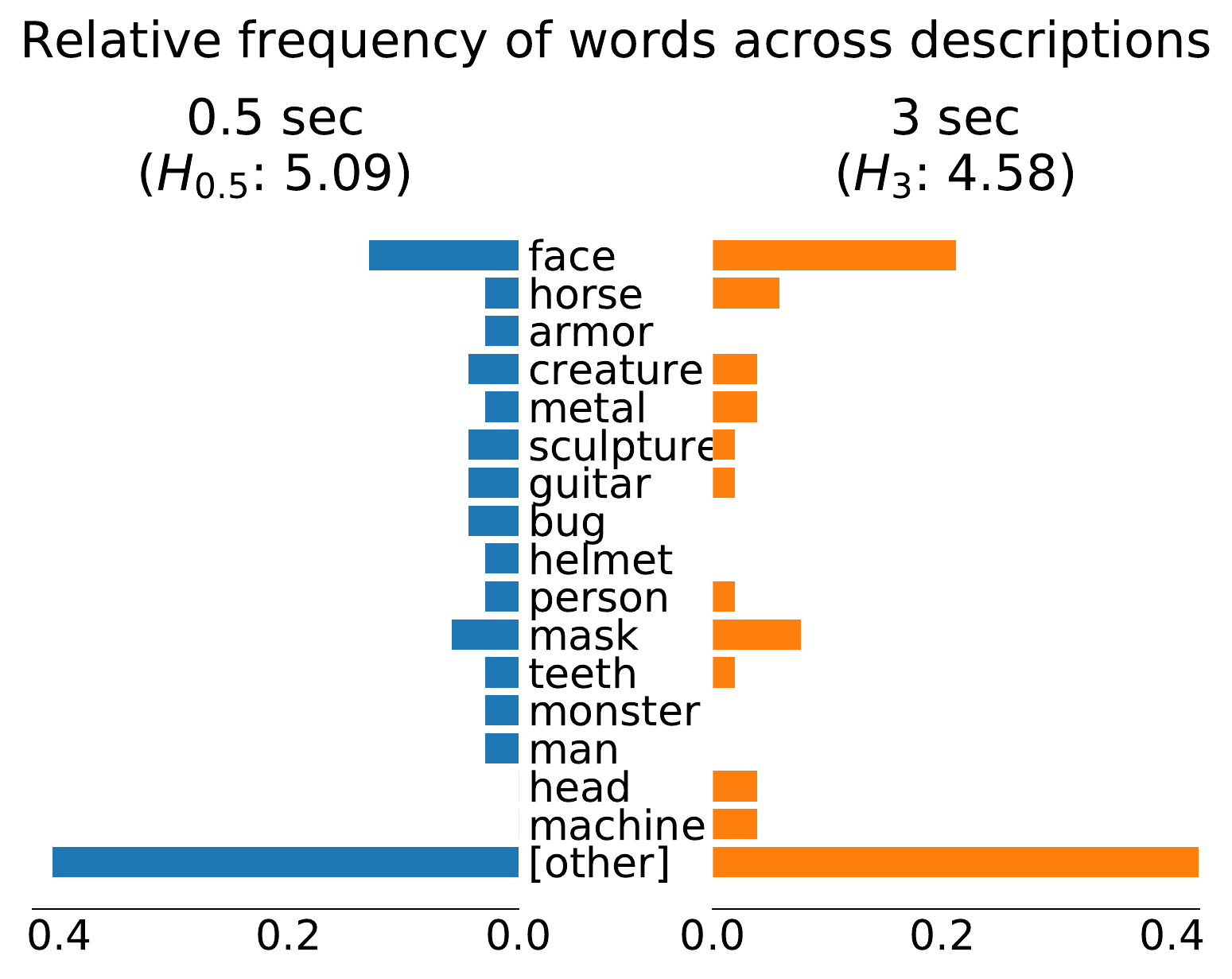}
\\

(a) & (b) \\
\addlinespace[0.3cm]

\includegraphics[width=\hwfigwidthtwo]{bf60de07a52b9e44b5e5.jpg}
\includegraphics[width=\hwhistwidthtwo]{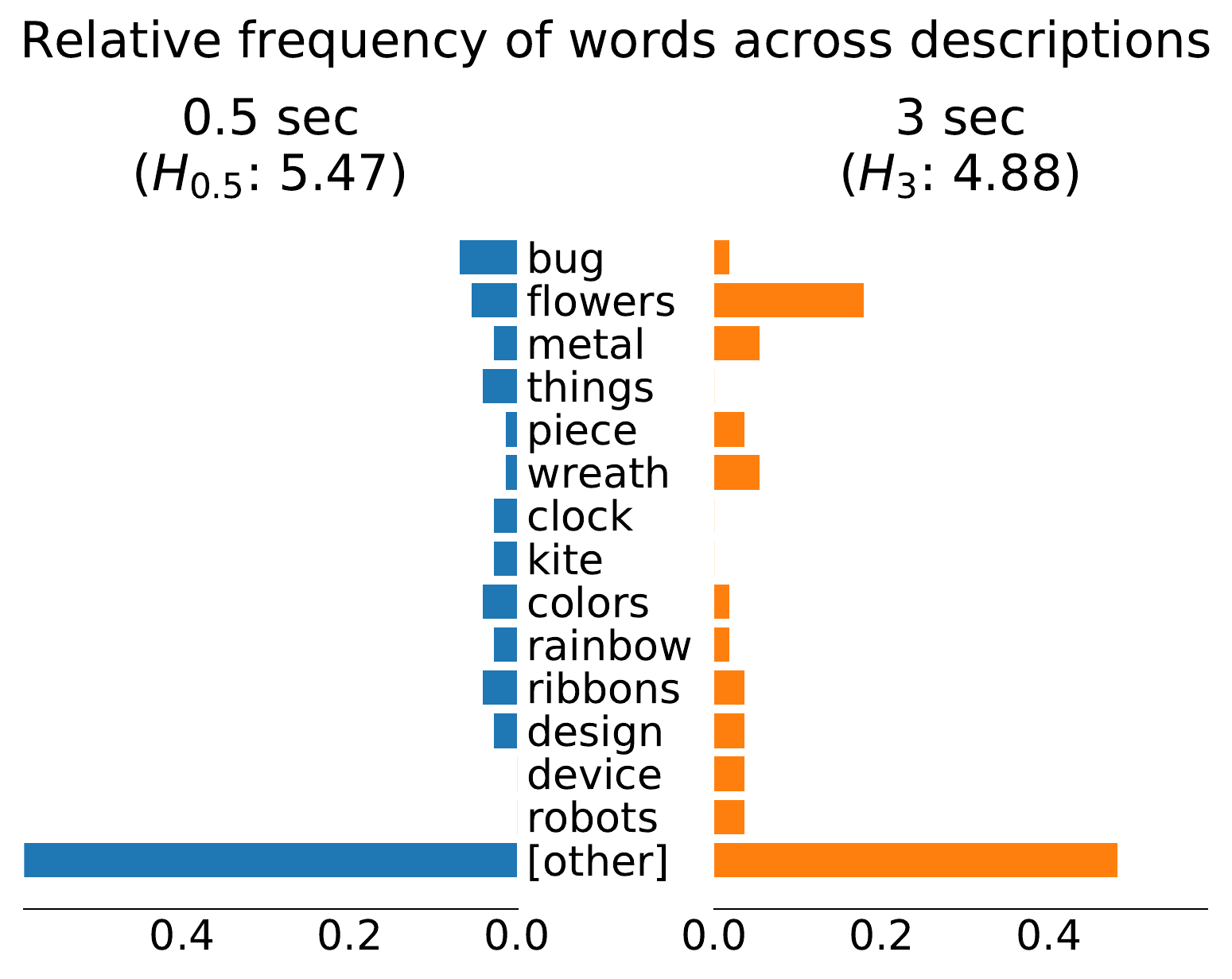}

&
\includegraphics[width=\hwfigwidthtwo]{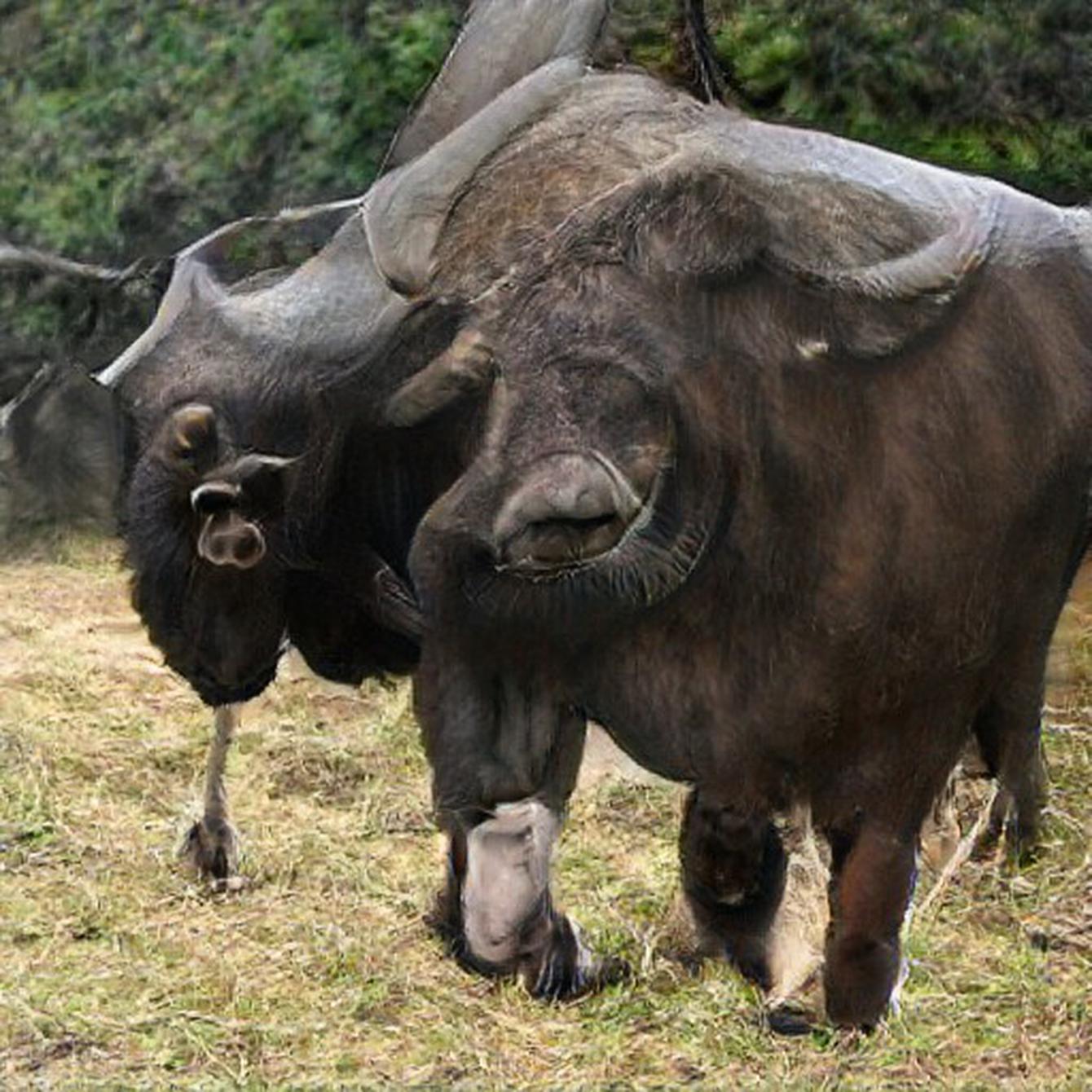}
\includegraphics[width=\hwhistwidthtwo]{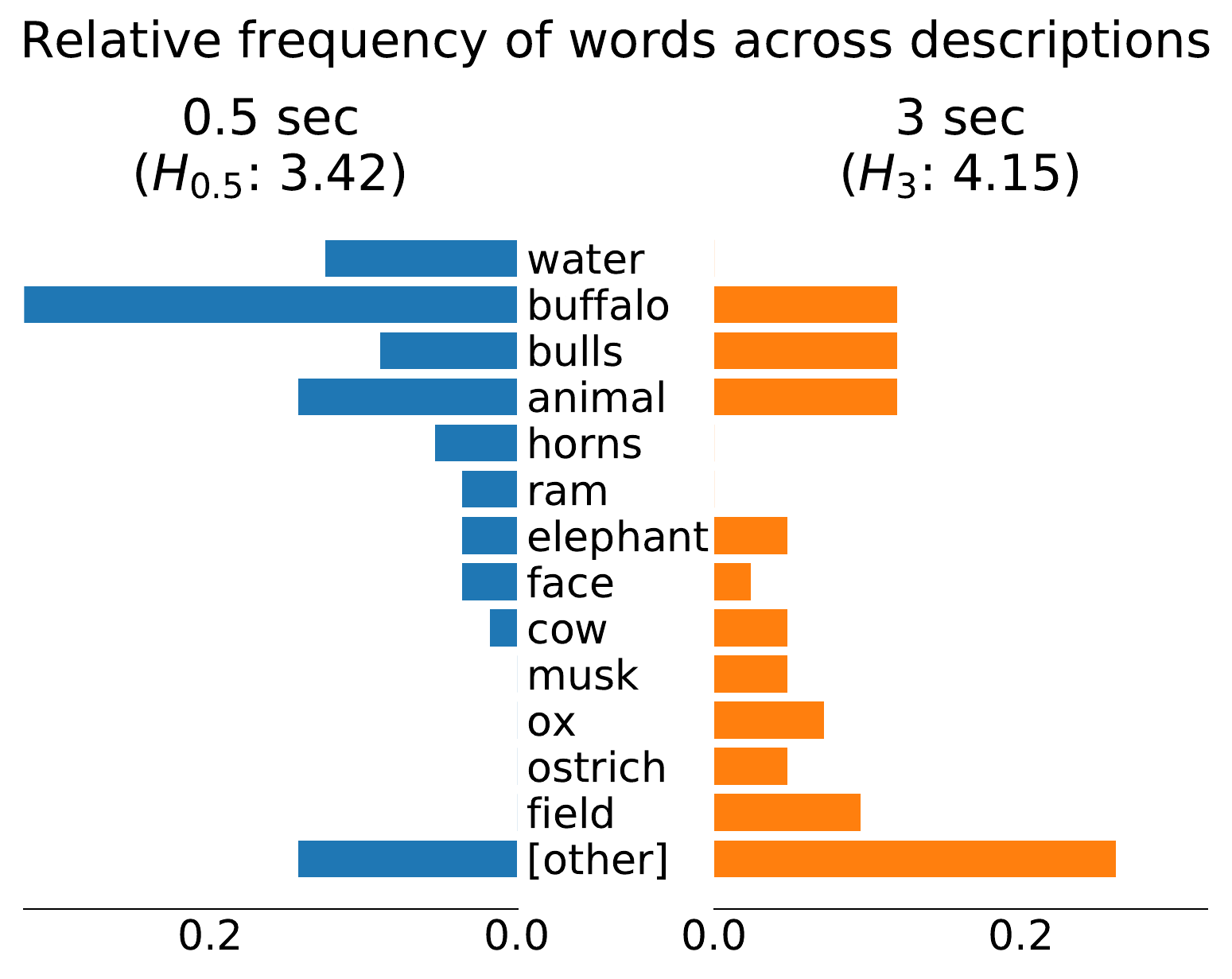}
\\

(c) & (d) \\

\addlinespace[0.3cm]

\includegraphics[width=\hwfigwidthtwo]{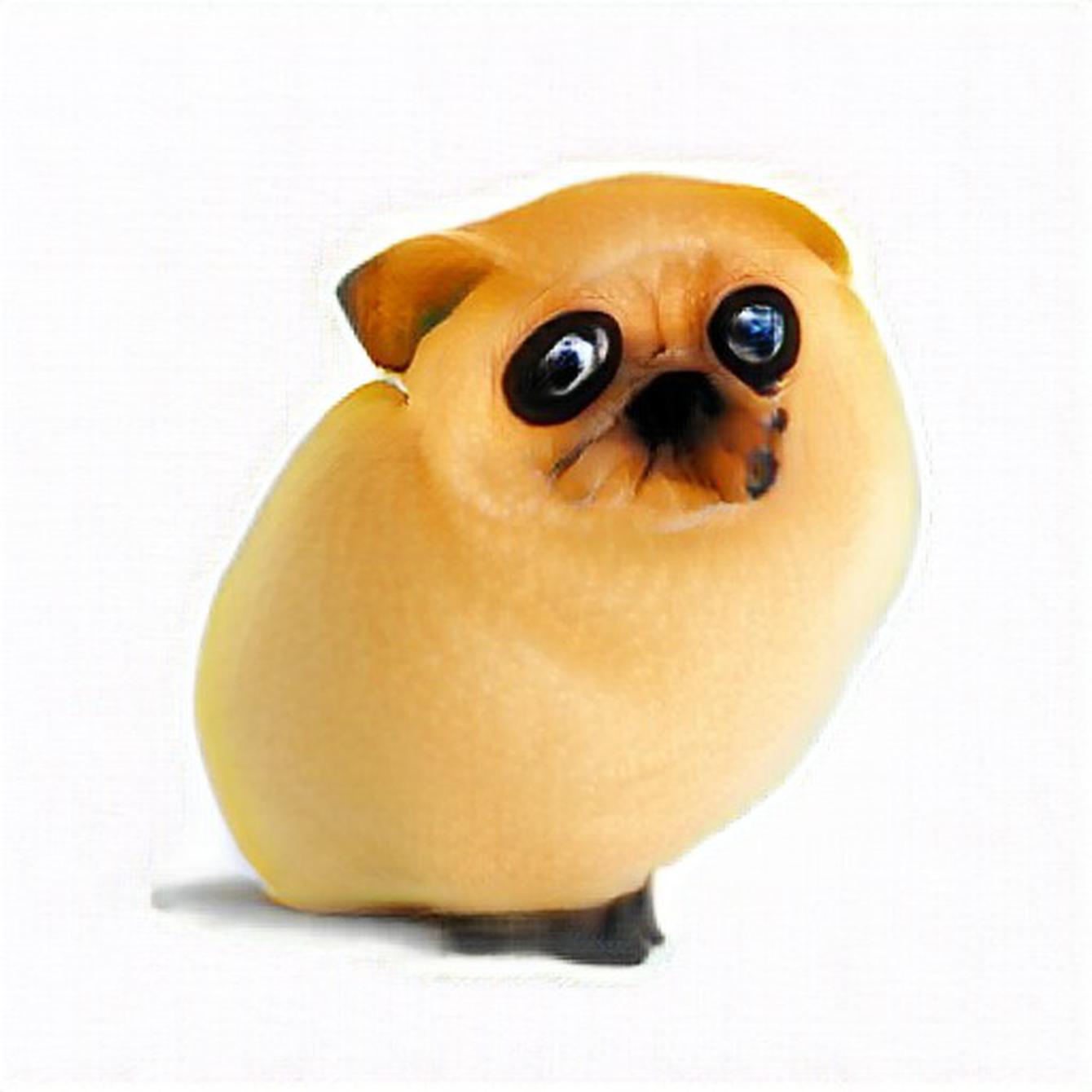}
\includegraphics[width=\hwhistwidthtwo]{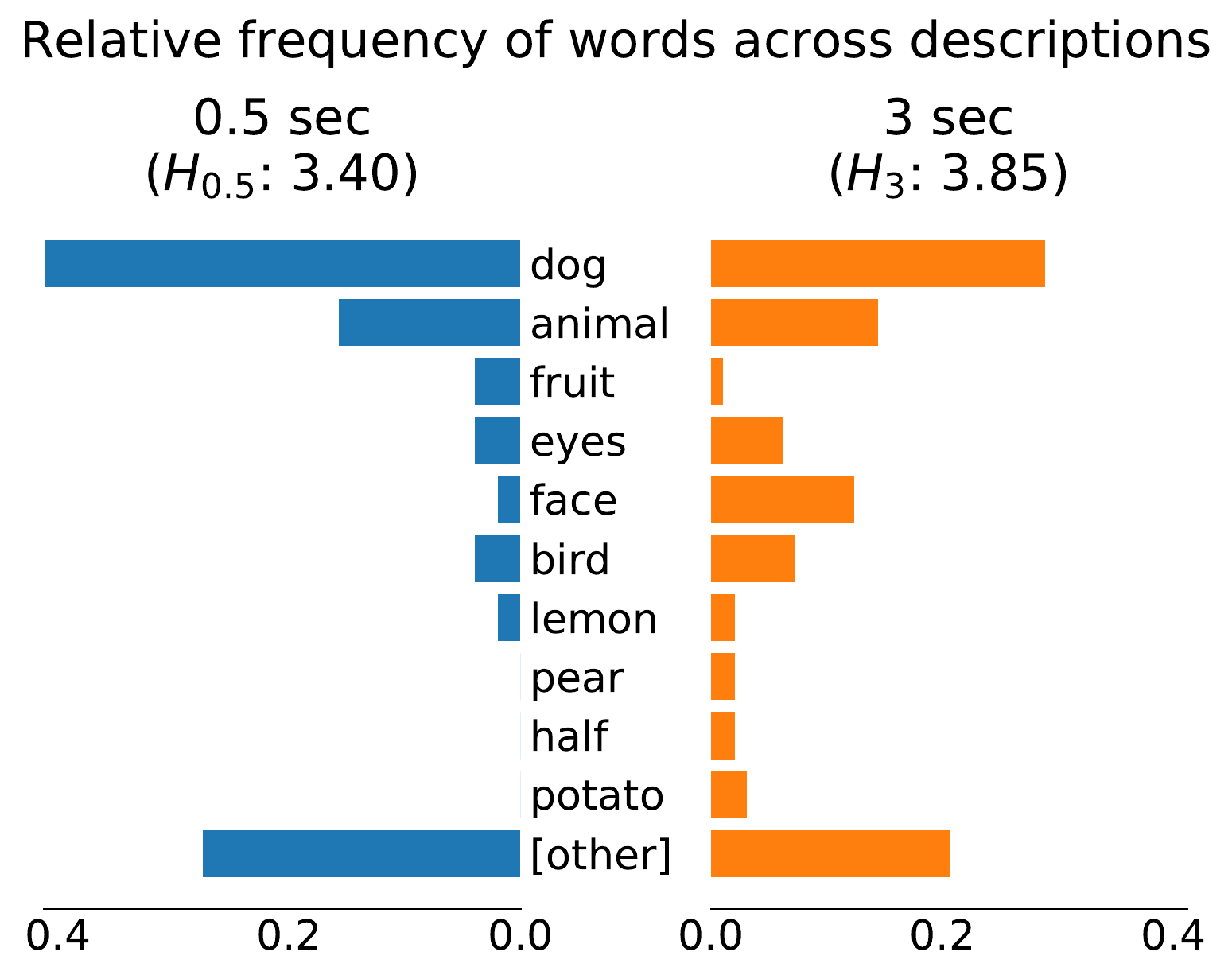}

&
\includegraphics[width=\hwfigwidthtwo]{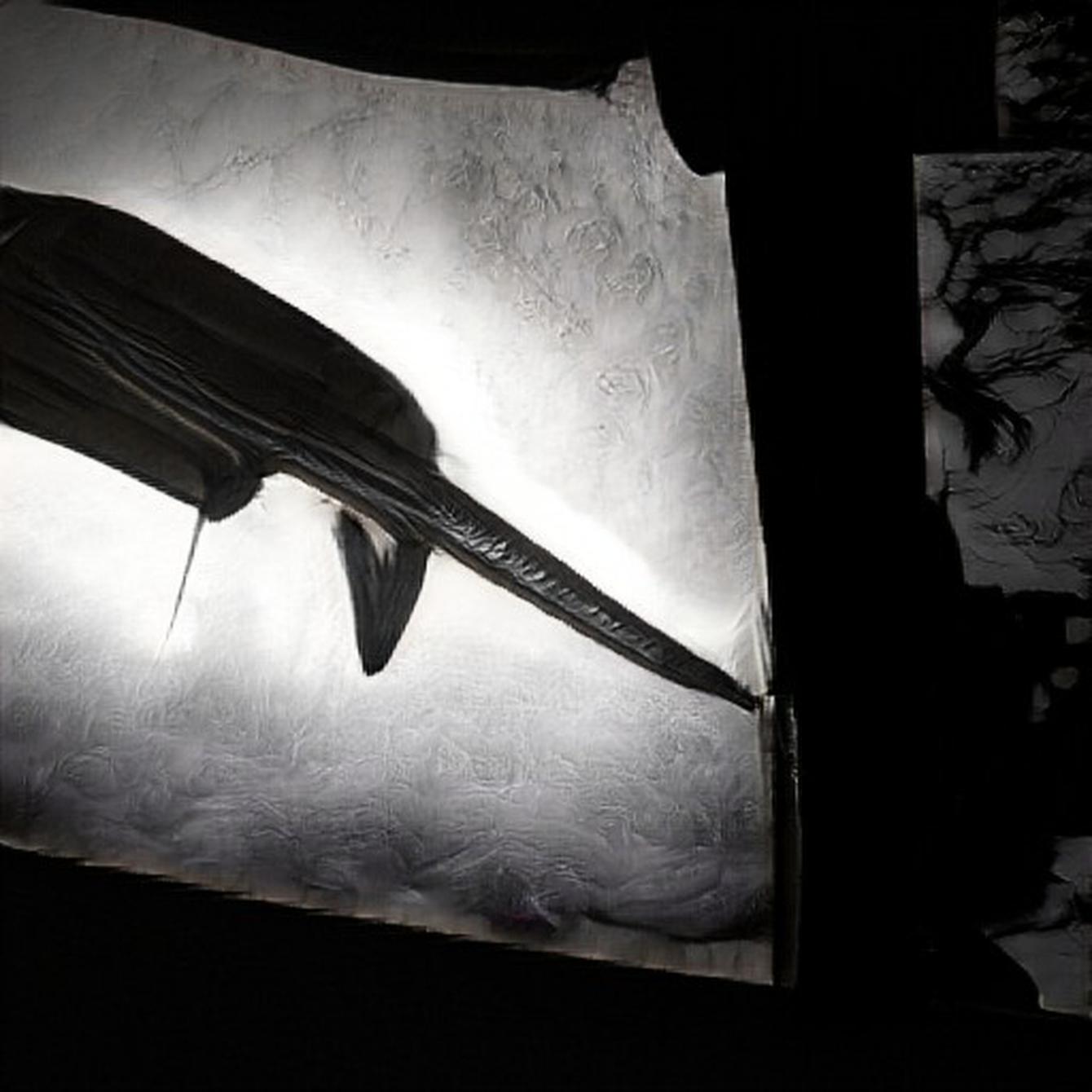}
\includegraphics[width=\hwhistwidthtwo]{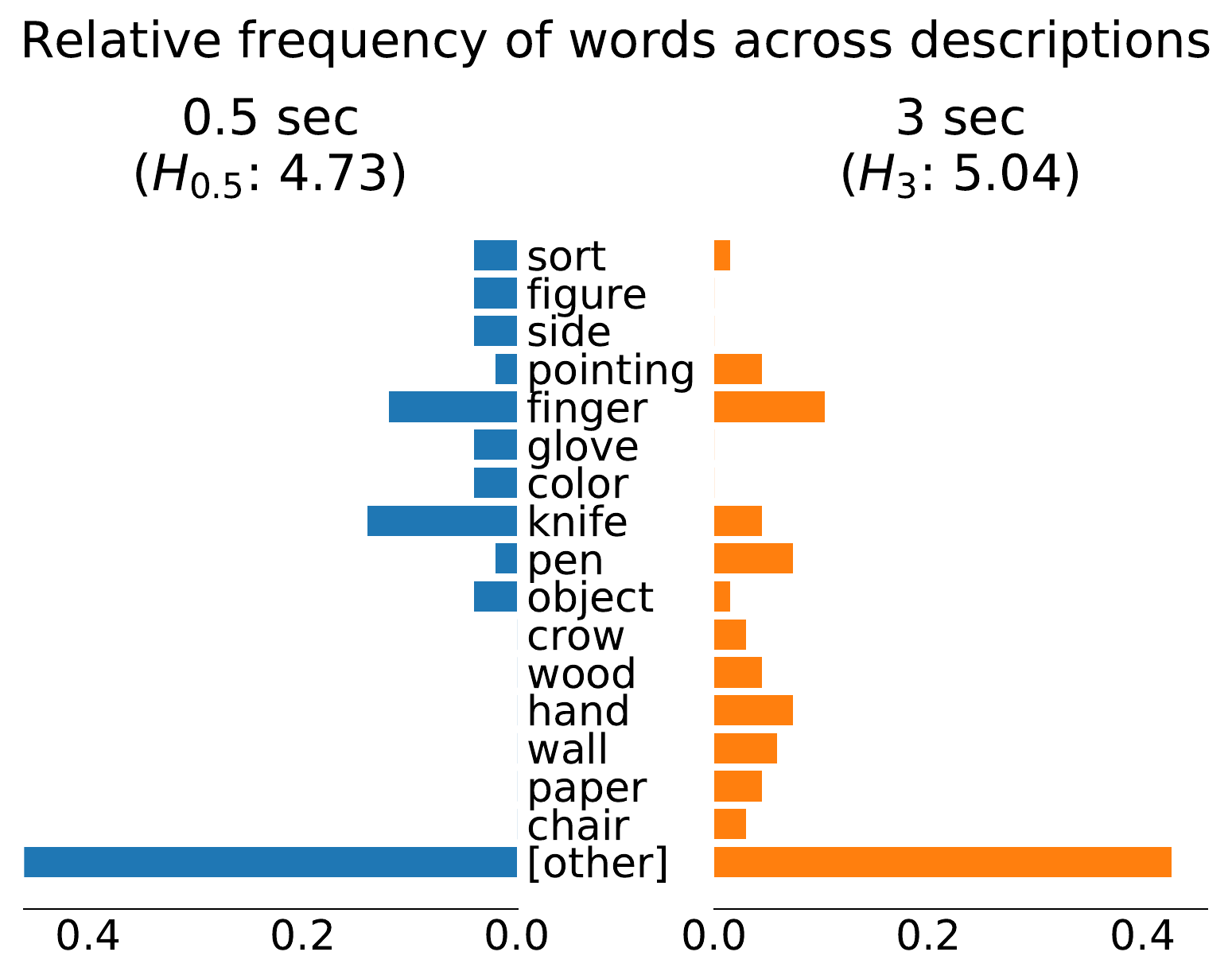}

\\

(e) & (f) 

\\ 

\addlinespace[0.3cm]
\includegraphics[width=\hwfigwidthtwo]{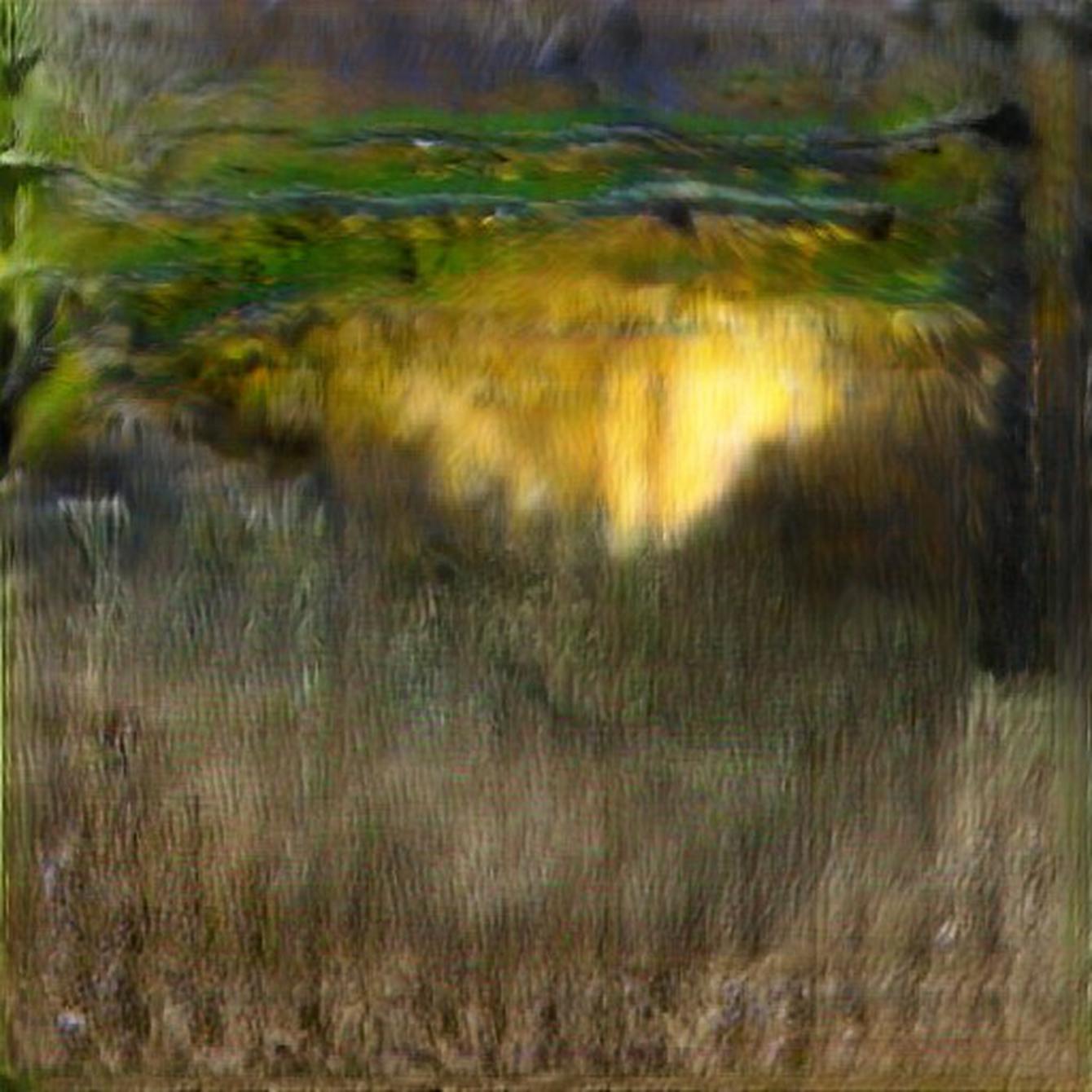}
\includegraphics[width=\hwhistwidthtwo]{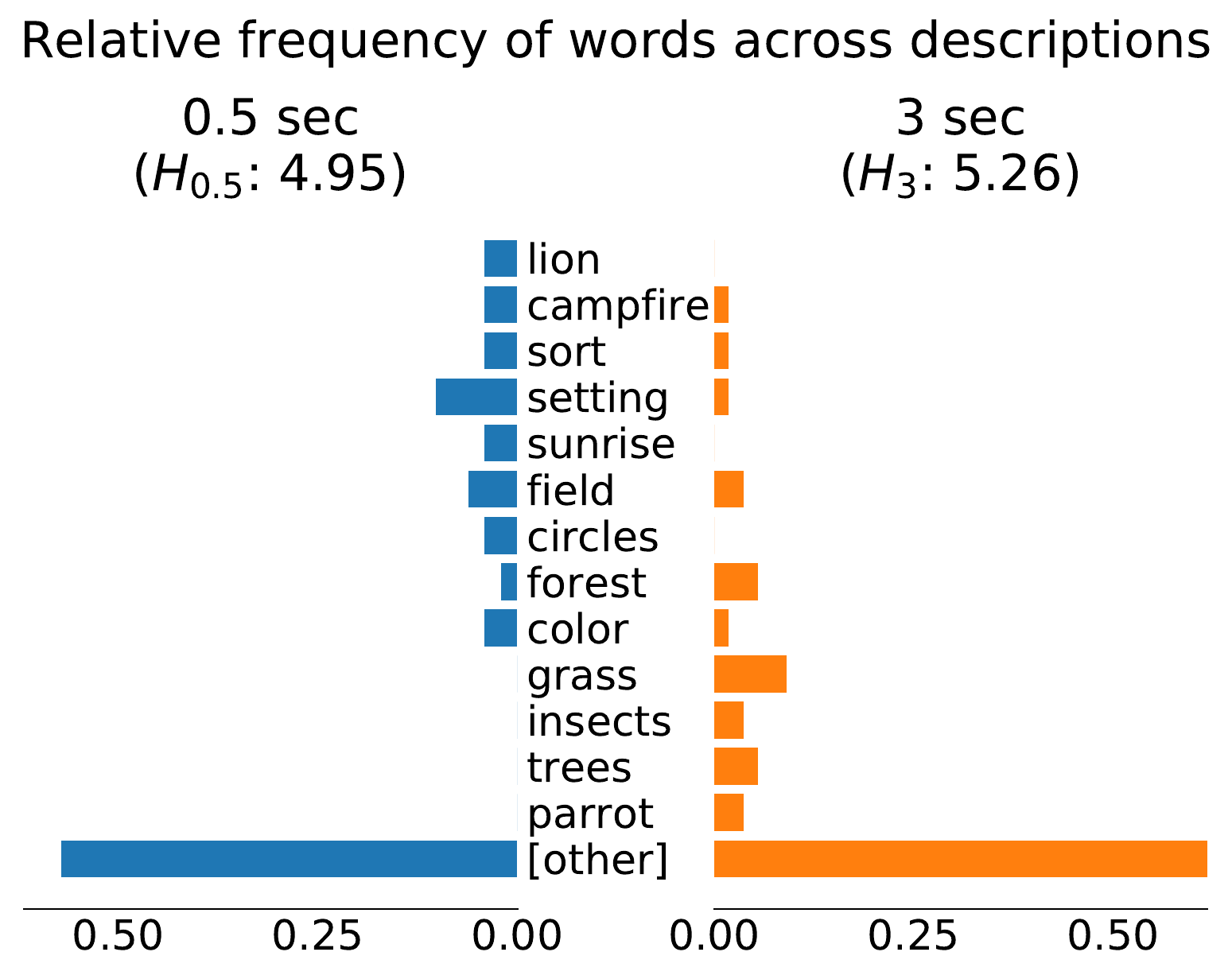}
&
\includegraphics[width=\hwfigwidthtwo]{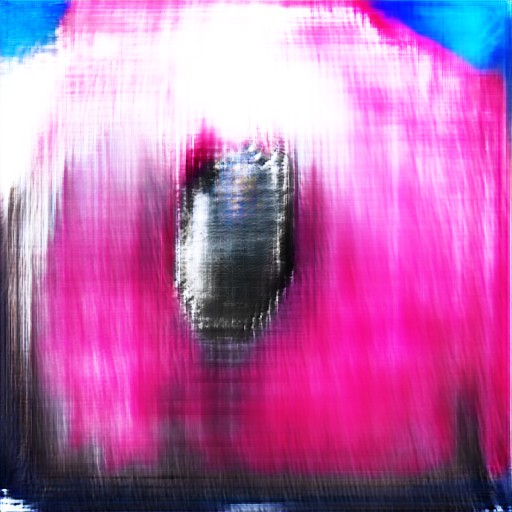}
\includegraphics[width=\hwhistwidthtwo]{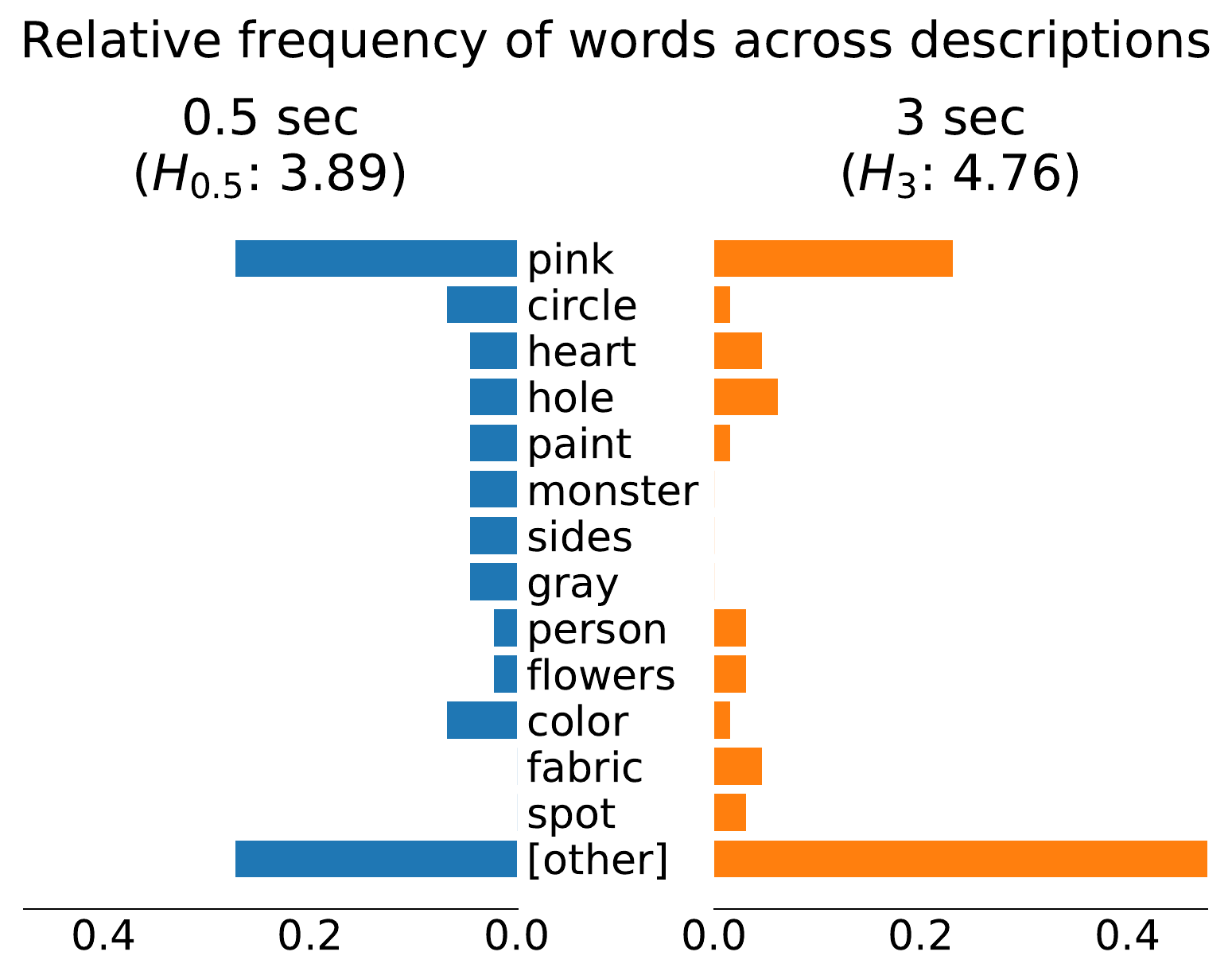}

\\
(g) & (h)
\end{tabular}

\caption{More interesting examples of high-entropy images. See text for discussion.
}
\label{fig:morefigures}
\end{figure*}

In Figure \ref{fig:morefigures}(d,e,f), entropy increases: viewers first give consistent ``first impression'' descriptions, such as ``buffalo'', ``dog'' and ``knife'', but the descriptions become considerably more varied and diverging when viewers have spent more time viewing them.

The final images are two examples that could be seen as purely abstract art, yet viewers can 
\newtext{occasionally perceive objects within them, though less consistently. Most tokens appear only once, as indicated by the large ``[other]'' category.}

\section{Discussion}

Our work is the first  to attempt to go beyond the high-level impressions of ambiguity by viewers, to instead uncover and quantify how ambiguous an image is across a population of viewers. \newtext{It is partly motivated by a desire to investigate whether modern computational methods can be used to address longstanding questions about subjective responses to images, particularly in light of the claims made by artists, such as Picasso, about what makes certain images aesthetically valuable.}

As we show, the \newtext{textual} response histograms generated by our method robustly capture image properties like indeterminacy, \newtext{measuring} shades of ambiguity which previous studies were \newtext{insensitive} to.

\newtext{Our preliminary study is intended to illustrate the potential value of this approach for measuring the subtle and highly subjective phenomenon of visual ambiguity. There are several limitations and directions for future development.
First, we are yet to validate the indeterminacy hypothesis suggested by the art theoretical literature, that there is a positive correlation between image ambiguity and aesthetic appreciation. The tools and methods developed here provide a useful starting point from which to rigorously investigate this claim in future.} 

\newtext{Second, on the basis of previous literature, we expected to find an increase in entropy over time for images with high levels of indeterminacy. We did not see this specific effect in our results (Figures \ref{fig:entropy_ranking}--\ref{fig:difference_ranking_comp}), but we did find change in entropy to provide some differentiation between types of ambiguity.
In fact, our results suggest that entropy alone can measure indeterminacy: if one wishes to follow Picasso's goal of producing the most associations, an image should not be realistic, but it should also not be too abstract.}

\newtext{Third, our text processing procedure is very simple. As a result, it discards some important information in the textual descriptions}, e.g., it cannot distinguish between an image having diverse descriptions because the image is confusing \newtext{(``it's either a horse or a chair'')}, because it is dichotomous \newtext{(``it's a chair made to look like a horse'')}, because it is complex \newtext{(``it's a horse next to a chair''), or because the description is  verbose (``it's a chair sitting on the ground'')}.  
\newtext{Our heuristic list of disallowed words could be replaced with a more nuanced filtering.}
\newtext{It is also unclear how best to take advantage of semantic similarity between descriptions that use distinct but related words.}
\newtext{Our text processing method ignored phrases that indicated difficulty or inability to respond, such as ``I'm not sure but...'' or ``I have no idea'', which occurred frequently for images in our indeterminate categories; the frequency of such phrases could be used to further inform the analysis}. 
Application of more sophisticated text processing \newtext{can} yield \newtext{more reliable and finer-grained} insights. 


\newtext{Another important future step is to relate entropies to aesthetic properties.  If semantic diversity and uncertainty are regarded as positive aesthetic attributes in artworks, as the art historical literature suggests, then we might expect to find a correlation between these qualities and entropy. Indeed, previous researchers have found such a positive effect \cite{jakesch}. As a preliminary test, we asked a separate sample of 128 crowdworkers to rate the images according to their level of ``interestingness'', ``powerfulness'', and ``engagement''; however, we found that crowdworkers gave highest scores to realistic, easily-interpretable images.} 
\newtext{There may be many reasons for this finding that do not necessarily invalidate the main hypothesis of this study, including viewer expectations for which images  constitute art images versus non-art images, the framing and setup of the task, and the expertise of the raters, each of which may need to be controlled for in future experiments.}
%


\newtext{The exposure times used in the present study were relatively short, particularly in the context of art viewing where a museum-goer might typically spend at least several tens of seconds studying an artwork in order to appreciate its nuances, and may return to look at it more than once~\cite{carbon2017art}. Our crowdworkers often commented that they had too little time to complete their descriptions. Exposures in the order of tens of seconds or minutes may yield more complex and nuanced responses. }

We thus far have only studied ambiguity of object recognition, whereas other image properties like figure-ground segmentation may also be ambiguous. However, our methodology suggests a general approach for probing perceptual uncertainty that could be combined with methods for crowdsourcing other image properties \cite{koenderinkAmbiguity, labelme,Gingold:2012:MPH}, to obtain a rich analysis of image ambiguity.

A suitably rich model of image ambiguity \newtext{can open up exciting future avenues} for image synthesis \newtext{applications. For instance, such a model may be able to guide GANs and related image synthesis pipelines towards more  interesting and unexpected creations. It may be used to automatically curate or filter image streams, or as an alternative metric of diversity to score different image synthesis techniques.}

\newtext{Finally, the methods being developed here could provide a potentially powerful computational framework for studying topics of interest to psychologists and neuroscientists, such as image perception, object recognition, and associative memory. }



\section*{Acknowledgements}

We thank Aude Oliva for feedback and support for this project and for hosting X.W.~as a visiting student, and Joel Simon for providing data from Artbreeder. 
All stimuli are public domain imagery obtained from Artbreeder, created by the following users.
Figure~\ref{fig:pipeline}:
guidoheinze, kent4747;
Figure~\ref{fig:entropy_ranking}:
jakritger, caincaser, strangecircus;
Figure~\ref{fig:difference_ranking}: 
desleep, jeffgiddens, thunderdog, angrytree607, portjos, strangecircus;
Figure~\ref{fig:difference_ranking_comp}:
jakritger, happyemil, thelindamartinez06;
Figure~\ref{fig:morefigures}:
thunderdog, desleep, spihut, telmaroza.

\bibliographystyle{ACM-Reference-Format}
\bibliography{indeterminacy}


\begin{thebibliography}{37}


\ifx \showCODEN    \undefined \def \showCODEN     #1{\unskip}     \fi
\ifx \showDOI      \undefined \def \showDOI       #1{#1}\fi
\ifx \showISBNx    \undefined \def \showISBNx     #1{\unskip}     \fi
\ifx \showISBNxiii \undefined \def \showISBNxiii  #1{\unskip}     \fi
\ifx \showISSN     \undefined \def \showISSN      #1{\unskip}     \fi
\ifx \showLCCN     \undefined \def \showLCCN      #1{\unskip}     \fi
\ifx \shownote     \undefined \def \shownote      #1{#1}          \fi
\ifx \showarticletitle \undefined \def \showarticletitle #1{#1}   \fi
\ifx \showURL      \undefined \def \showURL       {\relax}        \fi
\providecommand\bibfield[2]{#2}
\providecommand\bibinfo[2]{#2}
\providecommand\natexlab[1]{#1}
\providecommand\showeprint[2][]{arXiv:#2}

\bibitem[\protect\citeauthoryear{Bird, Klein, and Loper}{Bird
  et~al\mbox{.}}{2009}]%
        {nltk}
\bibfield{author}{\bibinfo{person}{Steven Bird}, \bibinfo{person}{Ewan Klein},
  {and} \bibinfo{person}{Edward Loper}.} \bibinfo{year}{2009}\natexlab{}.
\newblock \bibinfo{booktitle}{\emph{Natural Language Processing with Python}}.
\newblock \bibinfo{publisher}{O’Reilly Media Inc}.
\newblock


\bibitem[\protect\citeauthoryear{Boroș, Dumitrescu, and Burtica}{Boroș
  et~al\mbox{.}}{2018}]%
        {boro-dumitrescu-burtica:2018:K18-2}
\bibfield{author}{\bibinfo{person}{Tiberiu Boroș},
  \bibinfo{person}{Stefan~Daniel Dumitrescu}, {and} \bibinfo{person}{Ruxandra
  Burtica}.} \bibinfo{year}{2018}\natexlab{}.
\newblock \showarticletitle{{NLP}-Cube: End-to-End Raw Text Processing With
  Neural Networks}. In \bibinfo{booktitle}{\emph{Proceedings of the {CoNLL}
  2018 Shared Task: Multilingual Parsing from Raw Text to Universal
  Dependencies}}. \bibinfo{publisher}{Association for Computational
  Linguistics}, \bibinfo{address}{Brussels, Belgium},
  \bibinfo{pages}{171--179}.
\newblock
\urldef\tempurl%
\url{http://www.aclweb.org/anthology/K18-2017}
\showURL{%
\tempurl}


\bibitem[\protect\citeauthoryear{Brady, Konkle, Alvarez, and Oliva}{Brady
  et~al\mbox{.}}{2008}]%
        {brady2008visual}
\bibfield{author}{\bibinfo{person}{Timothy~F Brady}, \bibinfo{person}{Talia
  Konkle}, \bibinfo{person}{George~A Alvarez}, {and} \bibinfo{person}{Aude
  Oliva}.} \bibinfo{year}{2008}\natexlab{}.
\newblock \showarticletitle{Visual long-term memory has a massive storage
  capacity for object details}.
\newblock \bibinfo{journal}{\emph{Proceedings of the National Academy of
  Sciences}} \bibinfo{volume}{105}, \bibinfo{number}{38}
  (\bibinfo{year}{2008}), \bibinfo{pages}{14325--14329}.
\newblock


\bibitem[\protect\citeauthoryear{Brady, Konkle, Gill, Oliva, and Alvarez}{Brady
  et~al\mbox{.}}{2013}]%
        {brady2013visual}
\bibfield{author}{\bibinfo{person}{Timothy~F Brady}, \bibinfo{person}{Talia
  Konkle}, \bibinfo{person}{Jonathan Gill}, \bibinfo{person}{Aude Oliva}, {and}
  \bibinfo{person}{George~A Alvarez}.} \bibinfo{year}{2013}\natexlab{}.
\newblock \showarticletitle{Visual long-term memory has the same limit on
  fidelity as visual working memory}.
\newblock \bibinfo{journal}{\emph{Psychological science}} \bibinfo{volume}{24},
  \bibinfo{number}{6} (\bibinfo{year}{2013}), \bibinfo{pages}{981--990}.
\newblock


\bibitem[\protect\citeauthoryear{Brock, Donahue, and Simonyan}{Brock
  et~al\mbox{.}}{2019}]%
        {biggan}
\bibfield{author}{\bibinfo{person}{Andrew Brock}, \bibinfo{person}{Jeff
  Donahue}, {and} \bibinfo{person}{Karen Simonyan}.}
  \bibinfo{year}{2019}\natexlab{}.
\newblock \showarticletitle{Large Scale GAN Training for High Fidelity Natural
  Image Synthesis}. In \bibinfo{booktitle}{\emph{Proc.~ICLR}}.
\newblock


\bibitem[\protect\citeauthoryear{Carbon}{Carbon}{2017}]%
        {carbon2017art}
\bibfield{author}{\bibinfo{person}{Claus-Christian Carbon}.}
  \bibinfo{year}{2017}\natexlab{}.
\newblock \showarticletitle{Art perception in the museum: How we spend time and
  space in art exhibitions}.
\newblock \bibinfo{journal}{\emph{i-Perception}} \bibinfo{volume}{8},
  \bibinfo{number}{1} (\bibinfo{year}{2017}),
  \bibinfo{pages}{2041669517694184}.
\newblock


\bibitem[\protect\citeauthoryear{Cowling}{Cowling}{2006}]%
        {picassoQuote}
\bibfield{author}{\bibinfo{person}{Elizabeth Cowling}.}
  \bibinfo{year}{2006}\natexlab{}.
\newblock \bibinfo{booktitle}{\emph{Visiting Picasso: The Notebooks and Letters
  of Roland Penrose}}.
\newblock \bibinfo{publisher}{Thames \& Hudson}.
\newblock


\bibitem[\protect\citeauthoryear{Daw and Courville}{Daw and Courville}{2007}]%
        {dawPigeon}
\bibfield{author}{\bibinfo{person}{Nathaniel~D. Daw} {and}
  \bibinfo{person}{Aaron~C. Courville}.} \bibinfo{year}{2007}\natexlab{}.
\newblock \showarticletitle{The Pigeon as Particle Filter}. In
  \bibinfo{booktitle}{\emph{Proc.~NIPS}}.
\newblock


\bibitem[\protect\citeauthoryear{Fairhall and Ishai}{Fairhall and
  Ishai}{2008}]%
        {fairhall2008}
\bibfield{author}{\bibinfo{person}{Scott~L. Fairhall} {and}
  \bibinfo{person}{Alumit Ishai}.} \bibinfo{year}{2008}\natexlab{}.
\newblock \showarticletitle{Neural correlates of object indeterminacy in art
  compositions}.
\newblock \bibinfo{journal}{\emph{Consciousness and Cognition}}
  \bibinfo{volume}{17} (\bibinfo{year}{2008}), \bibinfo{pages}{923--932}.
\newblock


\bibitem[\protect\citeauthoryear{Fei-Fei, Iyer, Koch, and Perona}{Fei-Fei
  et~al\mbox{.}}{2007}]%
        {feifeiJOV}
\bibfield{author}{\bibinfo{person}{Li Fei-Fei}, \bibinfo{person}{Asha Iyer},
  \bibinfo{person}{Christof Koch}, {and} \bibinfo{person}{Pietro Perona}.}
  \bibinfo{year}{2007}\natexlab{}.
\newblock \showarticletitle{{What do we perceive in a glance of a real-world
  scene?}}
\newblock \bibinfo{journal}{\emph{Journal of Vision}} \bibinfo{volume}{7},
  \bibinfo{number}{1} (\bibinfo{date}{01} \bibinfo{year}{2007}),
  \bibinfo{pages}{10--10}.
\newblock


\bibitem[\protect\citeauthoryear{Fosco, Newman, Sukhum, Zhang, Zhao, Oliva, and
  Bylinskii}{Fosco et~al\mbox{.}}{2020}]%
        {Fosco_2020_CVPR}
\bibfield{author}{\bibinfo{person}{Camilo Fosco}, \bibinfo{person}{Anelise
  Newman}, \bibinfo{person}{Pat Sukhum}, \bibinfo{person}{Yun~Bin Zhang},
  \bibinfo{person}{Nanxuan Zhao}, \bibinfo{person}{Aude Oliva}, {and}
  \bibinfo{person}{Zoya Bylinskii}.} \bibinfo{year}{2020}\natexlab{}.
\newblock \showarticletitle{How Much Time Do You Have? Modeling Multi-Duration
  Saliency}. In \bibinfo{booktitle}{\emph{The IEEE/CVF Conference on Computer
  Vision and Pattern Recognition (CVPR)}}.
\newblock


\bibitem[\protect\citeauthoryear{Gamboni}{Gamboni}{2004}]%
        {gamboni}
\bibfield{author}{\bibinfo{person}{Dario Gamboni}.}
  \bibinfo{year}{2004}\natexlab{}.
\newblock \bibinfo{booktitle}{\emph{Potential Images: Ambiguity and
  Indeterminacy in Modern Art}}.
\newblock \bibinfo{publisher}{Reaktion Books}.
\newblock


\bibitem[\protect\citeauthoryear{Gingold, Shamir, and Cohen-Or}{Gingold
  et~al\mbox{.}}{2012}]%
        {Gingold:2012:MPH}
\bibfield{author}{\bibinfo{person}{Yotam Gingold}, \bibinfo{person}{Ariel
  Shamir}, {and} \bibinfo{person}{Daniel Cohen-Or}.}
  \bibinfo{year}{2012}\natexlab{}.
\newblock \showarticletitle{Micro Perceptual Human Computation}.
\newblock \bibinfo{journal}{\emph{ACM Transactions on Graphics (TOG)}}
  \bibinfo{volume}{31}, \bibinfo{number}{5}, Article \bibinfo{articleno}{119}
  (\bibinfo{date}{Aug.} \bibinfo{year}{2012}), \bibinfo{numpages}{12}~pages.
\newblock
\urldef\tempurl%
\url{https://doi.org/10.1145/2231816.2231817}
\showDOI{\tempurl}


\bibitem[\protect\citeauthoryear{Goodfellow, Pouget-Abadie, Mirza, Xu,
  Warde-Farley, Ozair, Courville, and Bengio}{Goodfellow et~al\mbox{.}}{2014}]%
        {GANs}
\bibfield{author}{\bibinfo{person}{Ian~J. Goodfellow}, \bibinfo{person}{Jean
  Pouget-Abadie}, \bibinfo{person}{Mehdi Mirza}, \bibinfo{person}{Bing Xu},
  \bibinfo{person}{David Warde-Farley}, \bibinfo{person}{Sherjil Ozair},
  \bibinfo{person}{Aaron Courville}, {and} \bibinfo{person}{Yoshua Bengio}.}
  \bibinfo{year}{2014}\natexlab{}.
\newblock \showarticletitle{Generative {A}dversarial {N}ets}. In
  \bibinfo{booktitle}{\emph{Proc. Neural Information Processing Systems}}.
\newblock


\bibitem[\protect\citeauthoryear{Hertzmann}{Hertzmann}{2010}]%
        {HertzmannScienceOfArt}
\bibfield{author}{\bibinfo{person}{Aaron Hertzmann}.}
  \bibinfo{year}{2010}\natexlab{}.
\newblock \showarticletitle{Non-{P}hotorealistic {R}endering and the {S}cience
  of {A}rt}. In \bibinfo{booktitle}{\emph{Proc.~NPAR}}.
\newblock


\bibitem[\protect\citeauthoryear{Hertzmann}{Hertzmann}{2020}]%
        {HertzmannIndeterminacy}
\bibfield{author}{\bibinfo{person}{Aaron Hertzmann}.}
  \bibinfo{year}{2020}\natexlab{}.
\newblock \showarticletitle{Visual Indeterminacy in GAN Art}.
\newblock \bibinfo{journal}{\emph{Leonardo}} \bibinfo{volume}{53},
  \bibinfo{number}{4} (\bibinfo{year}{2020}).
\newblock


\bibitem[\protect\citeauthoryear{Hofmann}{Hofmann}{1999}]%
        {plsi}
\bibfield{author}{\bibinfo{person}{Thomas Hofmann}.}
  \bibinfo{year}{1999}\natexlab{}.
\newblock \showarticletitle{Probabilistic Latent Semantic Indexing}. In
  \bibinfo{booktitle}{\emph{Proc. SIGIR}}. \bibinfo{pages}{50–57}.
\newblock


\bibitem[\protect\citeauthoryear{Ishai, Fairhall, and Pepperell}{Ishai
  et~al\mbox{.}}{2007}]%
        {fairhall_memory}
\bibfield{author}{\bibinfo{person}{Alumit Ishai}, \bibinfo{person}{Scott~L.
  Fairhall}, {and} \bibinfo{person}{Robert Pepperell}.}
  \bibinfo{year}{2007}\natexlab{}.
\newblock \showarticletitle{Perception, memory and aesthetics of indeterminate
  art}.
\newblock \bibinfo{journal}{\emph{Brain Research Bulletin}}
  \bibinfo{volume}{73}, \bibinfo{number}{4--6} (\bibinfo{year}{2007}).
\newblock


\bibitem[\protect\citeauthoryear{Jakesch, Leder, and Forster}{Jakesch
  et~al\mbox{.}}{2013}]%
        {jakesch}
\bibfield{author}{\bibinfo{person}{Martina Jakesch}, \bibinfo{person}{Helmut
  Leder}, {and} \bibinfo{person}{Michael Forster}.}
  \bibinfo{year}{2013}\natexlab{}.
\newblock \showarticletitle{Image Ambiguity and Fluency}.
\newblock \bibinfo{journal}{\emph{PLoS ONE}} \bibinfo{volume}{8},
  \bibinfo{number}{9} (\bibinfo{year}{2013}).
\newblock


\bibitem[\protect\citeauthoryear{Koenderink, van Doorn, Kappers, and
  Todd}{Koenderink et~al\mbox{.}}{2001}]%
        {koenderinkAmbiguity}
\bibfield{author}{\bibinfo{person}{Jan~J Koenderink}, \bibinfo{person}{Andrea~J
  van Doorn}, \bibinfo{person}{Astrid M~L Kappers}, {and}
  \bibinfo{person}{James~T Todd}.} \bibinfo{year}{2001}\natexlab{}.
\newblock \showarticletitle{Ambiguity and the ‘Mental Eye’ in Pictorial
  Relief}.
\newblock \bibinfo{journal}{\emph{Perception}} \bibinfo{volume}{30},
  \bibinfo{number}{4} (\bibinfo{year}{2001}), \bibinfo{pages}{431--448}.
\newblock


\bibitem[\protect\citeauthoryear{Muth and Carbon}{Muth and Carbon}{2013}]%
        {MuthAha2013}
\bibfield{author}{\bibinfo{person}{Claudia Muth} {and}
  \bibinfo{person}{Claus-Christian Carbon}.} \bibinfo{year}{2013}\natexlab{}.
\newblock \showarticletitle{The Aesthetic Aha: On the pleasure of having
  insights into Gestalt}.
\newblock \bibinfo{journal}{\emph{Acta Psychologica}} \bibinfo{volume}{144},
  \bibinfo{number}{1} (\bibinfo{year}{2013}), \bibinfo{pages}{25--30}.
\newblock


\bibitem[\protect\citeauthoryear{Muth and Carbon}{Muth and Carbon}{2016}]%
        {SeIns}
\bibfield{author}{\bibinfo{person}{Claudia Muth} {and}
  \bibinfo{person}{Claus-Christian Carbon}.} \bibinfo{year}{2016}\natexlab{}.
\newblock \showarticletitle{SeIns: Semantic Instability in Art}.
\newblock \bibinfo{journal}{\emph{Art \& Perception}} \bibinfo{volume}{4},
  \bibinfo{number}{1--2} (\bibinfo{year}{2016}), \bibinfo{pages}{145--184}.
\newblock


\bibitem[\protect\citeauthoryear{Muth, Hesslinger, and Carbon}{Muth
  et~al\mbox{.}}{2018}]%
        {variants}
\bibfield{author}{\bibinfo{person}{Claudia Muth}, \bibinfo{person}{Vera
  Hesslinger}, {and} \bibinfo{person}{Claus-Christian Carbon}.}
  \bibinfo{year}{2018}\natexlab{}.
\newblock \showarticletitle{Variants of Semantic Instability (SeIns) in the
  arts. A classification study based on experiential reports}.
\newblock \bibinfo{journal}{\emph{Psychology of Aesthetics, Creativity, and the
  Arts}} \bibinfo{volume}{12}, \bibinfo{number}{1} (\bibinfo{year}{2018}).
\newblock


\bibitem[\protect\citeauthoryear{Muth, Hesslinger, and Carbon}{Muth
  et~al\mbox{.}}{2015}]%
        {muth_challenge}
\bibfield{author}{\bibinfo{person}{Claudia Muth}, \bibinfo{person}{Vera~M.
  Hesslinger}, {and} \bibinfo{person}{Claus-Christian Carbon}.}
  \bibinfo{year}{2015}\natexlab{}.
\newblock \showarticletitle{The appeal of challenge in the perception of art:
  How ambiguity, solvability of ambiguity, and the opportunity for insight
  affect appreciation}.
\newblock \bibinfo{journal}{\emph{Psychology of Aesthetics, Creativity, and the
  Arts}} \bibinfo{volume}{9}, \bibinfo{number}{3} (\bibinfo{year}{2015}).
\newblock


\bibitem[\protect\citeauthoryear{Muth, Raab, and Carbon}{Muth
  et~al\mbox{.}}{2016}]%
        {MuthAha}
\bibfield{author}{\bibinfo{person}{Claudia Muth}, \bibinfo{person}{Marius~H.
  Raab}, {and} \bibinfo{person}{Claus-Christian Carbon}.}
  \bibinfo{year}{2016}\natexlab{}.
\newblock \showarticletitle{Semantic Stability is More Pleasurable in Unstable
  Episodic Contexts. On the Relevance of Perceptual Challenge in Art
  Appreciation}.
\newblock \bibinfo{journal}{\emph{Frontiers in Human Neuroscience}}
  \bibinfo{volume}{10} (\bibinfo{year}{2016}).
\newblock
\showISSN{1662-5161}
\urldef\tempurl%
\url{https://doi.org/10.3389/fnhum.2016.00043}
\showDOI{\tempurl}


\bibitem[\protect\citeauthoryear{Oliva}{Oliva}{2009}]%
        {oliva2009visual}
\bibfield{author}{\bibinfo{person}{Aude Oliva}.}
  \bibinfo{year}{2009}\natexlab{}.
\newblock \showarticletitle{Visual scene perception}.
\newblock In \bibinfo{booktitle}{\emph{Encyclopedia of perception}}.
  \bibinfo{publisher}{SAGE Publications, Inc. Thousand Oaks, CA},
  \bibinfo{pages}{1112--1117}.
\newblock


\bibitem[\protect\citeauthoryear{Pepperell}{Pepperell}{2011}]%
        {PepperellVI}
\bibfield{author}{\bibinfo{person}{Robert Pepperell}.}
  \bibinfo{year}{2011}\natexlab{}.
\newblock \showarticletitle{Connecting art and the brain: An artist's
  perspective on visual indeterminacy}.
\newblock \bibinfo{journal}{\emph{Frontiers in Human Neuroscience}}
  \bibinfo{volume}{5} (\bibinfo{year}{2011}).
\newblock


\bibitem[\protect\citeauthoryear{Pepperell}{Pepperell}{2015}]%
        {Pepperell}
\bibfield{author}{\bibinfo{person}{Robert Pepperell}.}
  \bibinfo{year}{2015}\natexlab{}.
\newblock \showarticletitle{Artworks as dichotomous objects: implications for
  the scientific study of aesthetic experience}.
\newblock \bibinfo{journal}{\emph{Front. Hum. Neurosci.}}  \bibinfo{volume}{9}
  (\bibinfo{date}{June} \bibinfo{year}{2015}).
\newblock


\bibitem[\protect\citeauthoryear{Potter}{Potter}{1975}]%
        {potter1975meaning}
\bibfield{author}{\bibinfo{person}{Mary~C Potter}.}
  \bibinfo{year}{1975}\natexlab{}.
\newblock \showarticletitle{Meaning in visual search}.
\newblock \bibinfo{journal}{\emph{Science}} \bibinfo{volume}{187},
  \bibinfo{number}{4180} (\bibinfo{year}{1975}), \bibinfo{pages}{965--966}.
\newblock


\bibitem[\protect\citeauthoryear{Potter}{Potter}{1999}]%
        {potter1999understanding}
\bibfield{author}{\bibinfo{person}{Mary~C Potter}.}
  \bibinfo{year}{1999}\natexlab{}.
\newblock \showarticletitle{Understanding sentences and scenes: The role of
  conceptual short-term memory}.
\newblock \bibinfo{journal}{\emph{Fleeting memories: Cognition of brief visual
  stimuli}} (\bibinfo{year}{1999}), \bibinfo{pages}{13--46}.
\newblock


\bibitem[\protect\citeauthoryear{Richter, Elger, and Obrist}{Richter
  et~al\mbox{.}}{2009}]%
        {richter}
\bibfield{author}{\bibinfo{person}{Gerhard Richter}, \bibinfo{person}{Dietmar
  Elger}, {and} \bibinfo{person}{Hans~Ulrich Obrist}.}
  \bibinfo{year}{2009}\natexlab{}.
\newblock \bibinfo{booktitle}{\emph{Gerhard Richter --- Text: Writing,
  Interviews and Letters 1961--2007}}.
\newblock \bibinfo{publisher}{Thames \& Hudson}, \bibinfo{address}{London}.
\newblock


\bibitem[\protect\citeauthoryear{Russell, Torralba, Murphy, and
  Freeman}{Russell et~al\mbox{.}}{2007}]%
        {labelme}
\bibfield{author}{\bibinfo{person}{Bryan Russell}, \bibinfo{person}{Antonio
  Torralba}, \bibinfo{person}{Kevin Murphy}, {and} \bibinfo{person}{William~T.
  Freeman}.} \bibinfo{year}{2007}\natexlab{}.
\newblock \showarticletitle{LabelMe: a database and web-based tool for image
  annotation}. In \bibinfo{booktitle}{\emph{Proc. ICCV}}.
\newblock


\bibitem[\protect\citeauthoryear{Schyns and Oliva}{Schyns and Oliva}{1994}]%
        {SchynsOliva94}
\bibfield{author}{\bibinfo{person}{Philippe~G Schyns} {and}
  \bibinfo{person}{Aude Oliva}.} \bibinfo{year}{1994}\natexlab{}.
\newblock \showarticletitle{From blobs to boundary edges: Evidence for time-and
  spatial-scale-dependent scene recognition}.
\newblock \bibinfo{journal}{\emph{Psychological science}} \bibinfo{volume}{5},
  \bibinfo{number}{4} (\bibinfo{year}{1994}), \bibinfo{pages}{195--200}.
\newblock


\bibitem[\protect\citeauthoryear{Torralba}{Torralba}{2009}]%
        {torralba2009many}
\bibfield{author}{\bibinfo{person}{Antonio Torralba}.}
  \bibinfo{year}{2009}\natexlab{}.
\newblock \showarticletitle{How many pixels make an image?}
\newblock \bibinfo{journal}{\emph{Visual neuroscience}} \bibinfo{volume}{26},
  \bibinfo{number}{1} (\bibinfo{year}{2009}), \bibinfo{pages}{123--131}.
\newblock


\bibitem[\protect\citeauthoryear{{Van de Cruys} and Wagemans}{{Van de Cruys}
  and Wagemans}{2011}]%
        {VandeCruys}
\bibfield{author}{\bibinfo{person}{Sander {Van de Cruys}} {and}
  \bibinfo{person}{Johan Wagemans}.} \bibinfo{year}{2011}\natexlab{}.
\newblock \showarticletitle{Putting reward in art: A tentative prediction error
  account of visual art}.
\newblock \bibinfo{journal}{\emph{i-Perception}} \bibinfo{volume}{2},
  \bibinfo{number}{9} (\bibinfo{year}{2011}).
\newblock


\bibitem[\protect\citeauthoryear{Wallraven, Kaulard, K\"{u}rner, Pepperell, and
  B\"{u}lthoff}{Wallraven et~al\mbox{.}}{2007a}]%
        {wallravenBeholder}
\bibfield{author}{\bibinfo{person}{C. Wallraven}, \bibinfo{person}{K. Kaulard},
  \bibinfo{person}{C. K\"{u}rner}, \bibinfo{person}{R. Pepperell}, {and}
  \bibinfo{person}{H. B\"{u}lthoff}.} \bibinfo{year}{2007}\natexlab{a}.
\newblock \showarticletitle{In the Eye of the Beholder - Perception of
  Indeterminate Art}. In \bibinfo{booktitle}{\emph{Proceedings of the Third
  Eurographics Conference on Computational Aesthetics in Graphics,
  Visualization and Imaging}}. \bibinfo{pages}{121–128}.
\newblock


\bibitem[\protect\citeauthoryear{Wallraven, Kaulard, K\"{u}rner, Pepperell, and
  B\"{u}lthoff}{Wallraven et~al\mbox{.}}{2007b}]%
        {wallraven_sap}
\bibfield{author}{\bibinfo{person}{Christian Wallraven},
  \bibinfo{person}{Kathrin Kaulard}, \bibinfo{person}{Cora K\"{u}rner},
  \bibinfo{person}{Robert Pepperell}, {and} \bibinfo{person}{Heinrich~H.
  B\"{u}lthoff}.} \bibinfo{year}{2007}\natexlab{b}.
\newblock \showarticletitle{Psychophysics for Perception of (in)Determinate
  Art}. In \bibinfo{booktitle}{\emph{Proceedings of the 4th Symposium on
  Applied Perception in Graphics and Visualization}}.
  \bibinfo{pages}{115–122}.
\newblock


\end{thebibliography}

\end{document}